\def\paperlanguage{}
\pdfoutput=1 

\documentclass[letterpaper, 10 pt, conference]{ieeeconf}  

\usepackage{bm}
\usepackage{cite}
\usepackage{flushend}
\include{preamble}

\IEEEoverridecommandlockouts                              

\title{\LARGE \textbf
  {
    \switchlanguage%
    {%
      Imitation Learning with Additional Constraints on Motion Style\\using Parametric Bias
    }%
    {%
      Parametric Biasを用いた動作スタイル制約を追加可能な模倣学習
    }%
  }
}

\author{Kento Kawaharazuka$^{1}$, Yoichiro Kawamura$^{1}$, Kei Okada$^{1}$, and Masayuki Inaba$^{1}$
  \thanks{$^{1}$ The authors are with the Department of Mechano-Informatics, Graduate School of Information Science and Technology, The University of Tokyo, 7-3-1 Hongo, Bunkyo-ku, Tokyo, 113-8656, Japan.
    {\texttt\small [kawaharazuka, y-kawamura, k-okada, inaba]@jsk.t.u-tokyo.ac.jp}
  }
}

\begin{document}

\maketitle
\thispagestyle{empty}
\pagestyle{empty}

\begin{abstract}
  \switchlanguage%
  {%
    Imitation learning is one of the methods for reproducing human demonstration adaptively in robots.
    So far, it has been found that generalization ability of the imitation learning enables the robots to perform tasks adaptably in untrained environments.
    However, motion styles such as motion trajectory and the amount of force applied depend largely on the dataset of human demonstration, and settle down to an average motion style.
    In this study, we propose a method that adds parametric bias to the conventional imitation learning network and can add constraints to the motion style.
    By experiments using PR2 and the musculoskeletal humanoid MusashiLarm, we show that it is possible to perform tasks by changing its motion style as intended with constraints on joint velocity, muscle length velocity, and muscle tension.
  }%
  {%
    模倣学習は人間の教示動作をロボットにおいて再生させるための手段の一つである.
    これまで, 模倣学習の汎化性により訓練に使用していない状況においても高い性能でタスクを実行できることが分かりつつある.
    その動作軌道・動作時の力の入れ具合等の動作スタイルは人間の教示動作というデータセットに大きく依存し, その中でも平均的な動作スタイルへと落ち着く.
    本研究では, 通常の模倣学習のネットワークにParametric Biasを追加し, 動作スタイルに対して制約を追加可能な手法を提案する.
    PR2, 筋骨格構造を持つロボットMusashiLarmにおいて, 関節速度や筋張力に制約を加え, 動作を意図的に変化させて再生できることを示す.
  }%
\end{abstract}

\section{Introduction}\label{sec:introduction}
\switchlanguage%
{%
  Robots are expected to perform a variety of tasks similar to humans.
  For this purpose, various model-based methods \cite{galindo2008planning} have been developed, while recently, methods based on deep learning have been increasing, such as reinforcement learning \cite{zhang2015reinforcement} and learning-based model predictive control \cite{kawaharazuka2020regrasp}.
  Among them, imitation learning \cite{osa2018imitation} learns behaviors directly from human demonstrations and thus can efficiently imitate human skills.
  So far, several methods have been developed, including those using image information \cite{yang2017repeatable, zhang2018imitation}, and those combining force information \cite{sasagawa2020imitation}.
  A method with meta-learning \cite{finn2017imitation} and a method with inverse reinforcement learning \cite{ho2016gail} have also been proposed.

  In these imitation learning methods, the motion style such as the motion trajectory and the amount of force applied is determined depending on the dataset of human demonstrations.
  Usually, the imitated behavior tends to converge into an average of various human demonstrations when the imitation model is trained.
  However, there are cases in which we want to reproduce not only the average behavior but also the bias in the variation of human motion styles.
  For example, we may want to apply as little force as possible, to suppress the speed and acceleration, or to keep some joints as still as possible.
  Considering motion style makes it possible to change the behavior in a way that suits the current desired style.
  However, there are few methods which can control the motion style in imitation learning.

  We summarize the methods considering motion style in imitation learning.
  First, there are classical methods that do not use deep learning, such as Probabilistic Movement Primitives \cite{paraschos2013promps}, a method using Dynamic Bayesian Network \cite{eppner2009imitation}, and a method using Probabilistic Principle Component Analysis \cite{perico2019imitation}, which can take into account additional constraints such as motion speed and obstacles.
  These methods make a number of assumptions about the structure of sensory and control signals, so while they can imitate behaviors from a small number of demonstrations, they are not scalable to more multi-dimensional and multi-modal systems, such as those involving raw images and tactile sensors.
  In contrast to these methods, InfoGAIL \cite{li2017infogail}, OptionGAN \cite{henderson2018optiongan}, and MSRD \cite{chen2020heterogeneous} use reinforcement learning and inverse reinforcement learning mechanisms, and they are capable of adaptive imitation learning from raw visual information.
  InfoGAIL can embed differences in motion style in latent space using mutual information, OptionGAN can create policies for each motion style, and MSRD can create rewards for each motion style.
  On the other hand, since these motion styles are discrete values, they cannot be controlled by additional constraints.
  In addition, since reinforcement learning is used, they require online trials which could be costly in robotic applications.
}%
{%
  ロボットには人間と同様に様々なタスクをこなすことが求められている.
  そのために様々なmodel-basedな手法\cite{galindo2008planning}が開発されてきた一方, 最近では深層学習を用いた手法が増え, 強化学習\cite{zhang2015reinforcement}や予測モデル\cite{kawaharazuka2020regrasp}等を使った手法が一般的になりつつある.
  この中でも, 模倣学習\cite{osa2018imitation}は人間のデモンストレーションから直接動作を学習するため, 効率よく人間のスキルを真似することができる.
  これまで, 画像情報を使ったものや\cite{yang2017repeatable, zhang2018imitation}, 力情報を組み合わせたもの等が開発されてきている\cite{sasagawa2020imitation}.
  また, meta-learningと組み合わせた研究\cite{finn2017imitation}や逆強化学習とGenerative Adversarial Networkを組み合わせた模倣学習手法\cite{ho2016gail}等も開発されている.

  この模倣学習は, 人間の教示に依存して動作軌道や力の入れ具合等の動作スタイルが決定される.
  バラついた人間の教示の中でも, 通常はその平均的な動作が再生されることになる.
  しかし, 単純に平均的な動作を行うのみではなく, 人間の動作のバラつきの中の偏りを再現したい場合がある.
  例えば, なるべく力がかからないようにしたり, 速度・加速度を抑えたり, 一部の関節をなるべく動かないようにしたり等である.
  これにより, その状況に合った形で振る舞いを変更していくことが可能となると考える.

  motion styleの違いを模倣学習の中で考慮した手法とその問題点をまとめる.
  まず, Probabilistic Movement Primitives \cite{paraschos2013promps}やDynamic Bayesian Networkを用いた手法\cite{eppner2009imitation}, Probabilistic Principle Component Analysis を用いた手法\cite{perico2019imitation}等のような, 深層学習を用いない古典的な手法があり, 動作速度や障害物等の追加制約を考慮可能である.
  これらは感覚や制御入力の構造にある多くの仮定を置くため, 少数のデモンストレーションから動作学習が可能である一方, 生の画像や接触覚を含むような, より多次元でmulti-modalな系にスケールしないという問題点がある.
  これらに対して, InfoGAIL \cite{li2017infogail}やOptionGAN \cite{henderson2018optiongan}, MSRD \cite{chen2020heterogeneous}等は強化学習・逆強化学習の仕組みを用いて, 生の視覚情報を扱った適応的な模倣学習が可能である.
  InfoGAILは相互情報量を使ってlatent spaceにmotion styleの違いを埋め込むことができ, OptionGANはstyleごとにpolicyを, MSRDはstyleごとにrewardを作成することで, motion styleを学習可能としている.
  一方, それらmotion styleは離散的な値であるため, そのmotion styleを追加制約をもとに制御することはできない.
  また, 強化学習を併用するため, demonstrationだけから模倣動作は学習できず, 試行錯誤をしながら最適なpolicyを獲得していく必要がある.
  その他にも, 模倣学習ではないものの, motion style transferの文脈でmotion styleを変化させることが行われてきている\cite{aberman2020unpaired}.
}%

\begin{figure}[t]
  \centering
  \includegraphics[width=1.0\columnwidth]{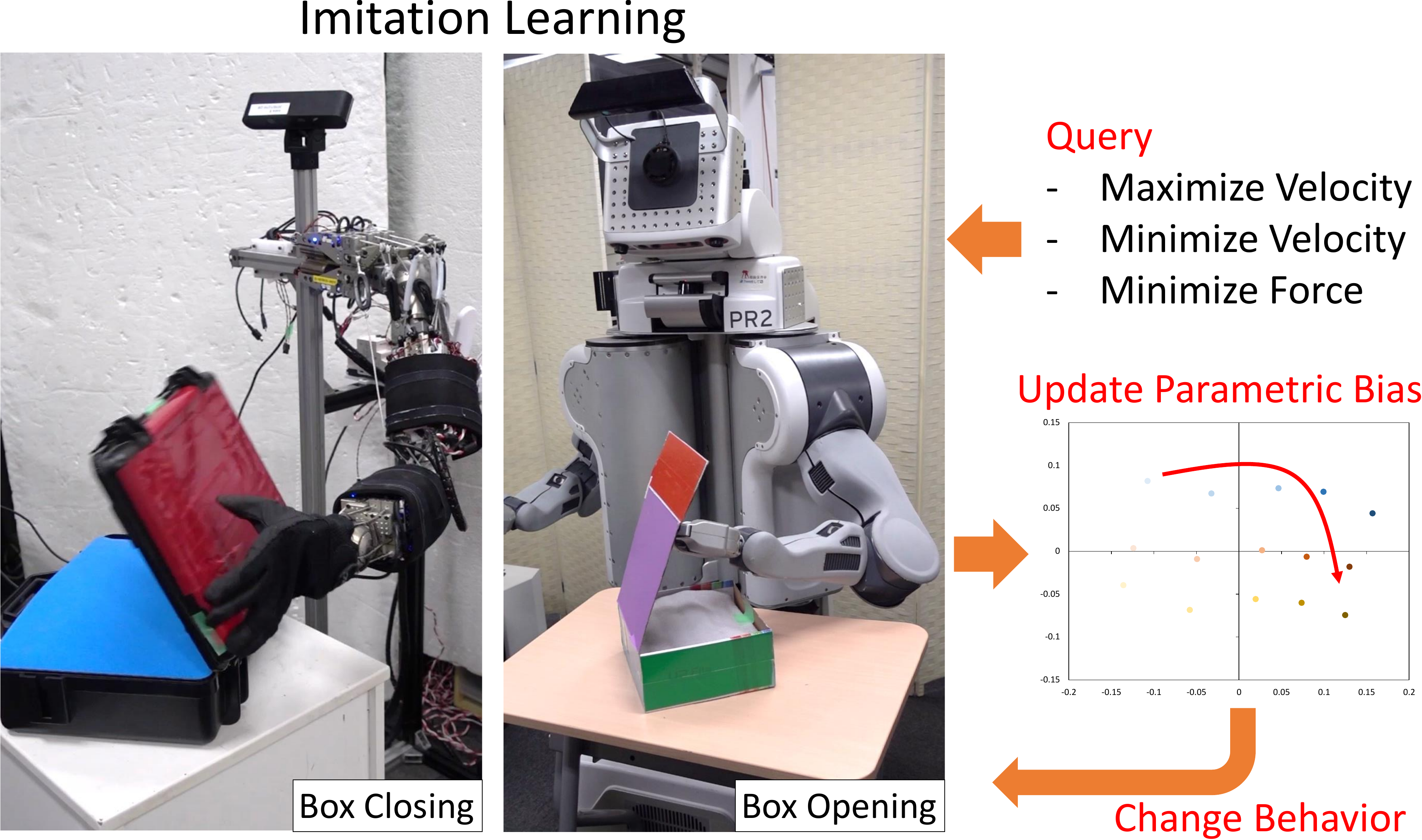}
  \vspace{-3.0ex}
  \caption{Imitation learning with additional constraints on motion style.}
  \label{figure:concept}
  \vspace{-3.0ex}
\end{figure}

\begin{figure*}[t]
  \centering
  \includegraphics[width=1.8\columnwidth]{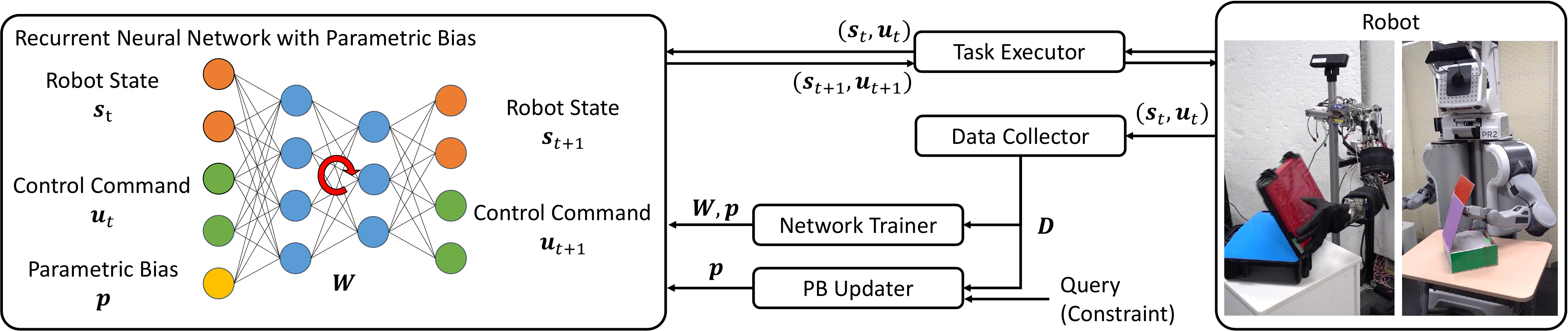}
  \vspace{-1.0ex}
  \caption{The overall system of imitation learning with additional constraints.}
  \label{figure:whole-system}
  \vspace{-3.0ex}
\end{figure*}

\switchlanguage%
{%
  In this study, we propose an imitation learning method that can control motion style using parametric bias \cite{tani2002parametric} (\figref{figure:concept}).
  So far, parametric bias has been used for the purpose of extracting multiple attractor dynamics.
  These methods have been used to extract the dynamics of objects from multi-modal sensors \cite{ogata2005rnnpb}, to extract the differences in the dynamics of object manipulation \cite{kawaharazuka2020dynamics}, and to extract the changes in the dynamics of the end effector due to tool grasping \cite{nishide2009toolbody}.
  In the imitation learning process, once the multiple attractor dynamics have been trained, the human shows a robot a behavior to be imitated, the robot determines the attractor dynamics to match it, and reproduces the imitation behavior.
  Based on these research, we develop a novel method that considers additional constraints on motion style, such as motion trajectory and the amount of force applied, instead of just changing the attractor dynamics to imitate the shown behaviors.
  Note that the word ``constraint'' in this study is used as soft constraint defined by loss function, as opposed to hard constraint.
  By including parametric bias in common imitation learning methods with recurrent neural network \cite{osa2018imitation}, embedding multiple motion styles in parametric bias allows us to freely control the motion style.
  We validate the effectiveness of this study by experiments using a simulated 1-DOF tendon arm, the musculoskeletal humanoid MusashiLarm \cite{kawaharazuka2019musashi}, and PR2.

  This study is organized as follows.
  In \secref{sec:proposed-method}, we describe the network structure of the recurrent neural network with parametric bias, its training, task execution, and update of parametric bias considering additional constraints on motion style.
  In \secref{sec:experiments}, we evaluate the basic performance by a simulated 1-DOF tendon arm, and perform box opening by PR2 and box closing by a musculoskeletal humanoid MusashiLarm.
  Finally, the experimental results are discussed in \secref{sec:discussion}, and the conclusion is given in \secref{sec:conclusion}.
}%
{%
  そこで本研究では, parametric bias \cite{tani2002parametric}を用いて追加制約を考慮可能な模倣学習手法を提案する(\figref{figure:concept}).
  これまでparametric biasは複数のattractor dynamicsを抽出する目的で使われてきた.
  multi-modalな感覚から物体のdynamicsを抽出したり\cite{ogata2005rnnpb}, 物体操作のdynamicsの違いを抽出したり\cite{yokoya2006rnnpb, kawaharazuka2020dynamics}, 道具把持による手先のdynamicsの変化を抽出している\cite{nishide2009toolbody}.
  模倣学習においては, それら複数のattractor dynamicsが訓練された後は, 一度人間が動作を見せたり道具画像等を見せたりすることで, それに合致するようにattractor dynamicsを決定し, これを元に模倣動作を再生する.
  これらの研究に基づき, これらのようにただ模倣すべき見せられた状態を模倣できるようにattractor dynamicsを変化させるのではなく, これに動作軌道や力の入れ具合等に関する追加の動作スタイル制約を考慮可能にする手法を開発する.
  通常のRNNを用いた模倣学習にparametric biasを追加し, 複数のmotion styleをこのparametric biasに埋め込むことで, motion styleを自在にcontrolすることが可能となる.
  本研究ではこのimitation learningを1自由度腱駆動シミュレータ, 筋骨格ヒューマノイドMusashiLarm \cite{kawaharazuka2019musashi}, PR2により検証し, その有効性を示す.

  本研究の構成は以下のようになっている.
  \secref{sec:proposed-method}では追加の動作スタイル制約を考慮可能な模倣学習について, ネットワーク構造・その学習・動作実行とparametric biasの更新について述べる.
  \secref{sec:experiments}では, 1自由度腱駆動シミュレータによる基本的な性能評価, PR2による箱開け動作, 筋骨格ヒューマノイドMusashiLarmによる箱閉め動作実験を行う.
  最後に, \secref{sec:discussion}で実験結果について議論し, \secref{sec:conclusion}で結論を述べる.
}%

\section{Imitation Learning with Additional Constraints on Motion Style} \label{sec:proposed-method}
\switchlanguage%
{%
  We show the overall system in \figref{figure:whole-system}.
}%
{%
  本研究の全体システムを\figref{figure:whole-system}に示す.
}%
%

\subsection{Network Structure} \label{subsec:network-structure}
\switchlanguage%
{%
  The recurrent neural network with parametric bias (RNNPB) used in this study, can be expressed as follows.
  The basic network structure is the same as \cite{tani2002parametric}.
  \begin{align}
    (\bm{s}_{t+1}, \bm{u}_{t+1}) = \bm{h}(\bm{s}_{t}, \bm{u}_{t}, \bm{p}) \label{eq:rnnpb}
  \end{align}
  where $t$ is the current time step, $\bm{s}$ is the sensor state, $\bm{u}$ is the control command, $\bm{p}$ is the parametric bias, and $\bm{h}$ is the function of RNNPB.
  $\bm{s}$ and $\bm{u}$ vary from robot to robot.
  For axis-driven robots, $\bm{u}$ is typically the joint angle command $\bm{\theta}^{ref}$ or joint torque command $\bm{\tau}^{ref}$.
  Although imitation learning by a tendon-driven robot has not been done before, in this study, the muscle length command $\bm{l}^{ref}$ is used as $\bm{u}$.
  The robot state $\bm{s}$ can be an arbitrary combination of sensory input such as image from camera, joint torque $\bm{\tau}$, and muscle tension $\bm{f}$.
  In this study, we use $\bm{z}$ for a latent representation of camera image, which is extracted by AutoEncoder \cite{hinton2006reducing}.
  Parametric bias $\bm{p}$ is a learnable variable that is identical within a demonstration but different between different demonstrations.
  Each sensor value is normalized using all the obtained data and then is input to the network.

  In this study, RNNPB has 10 layers, consisting of 4 fully-connected layers, 2 LSTM \cite{hochreiter1997lstm} layers, and 4 fully-connected layers in order.
  The number of units is set to \{$N_u+N_s+N_p$, 500, 300, 100, 100 (number of units in LSTM), 100 (number of units in LSTM), 100, 300, 500, $N_u+N_s$\} (where $N_{\{u, s, p\}}$ is the dimension of $\{\bm{u}, \bm{s}, \bm{p}\}$).
  The activation function is hyperbolic tangent and the update rule is Adam \cite{kingma2015adam}.
  When compressing the image, for a $96\times96$ RGB image, the convolutional layer with kernel size 3 and stride 2 is applied five times, and the dimension is reduced to 256 and 12 by the fully connected layers, and then the image is restored by the fully connected layers and the deconvolutional layers in the same way.
  The batch normalization \cite{ioffe2015batchnorm} and the activation function of ReLU \cite{nair2010relu} is applied to all layers except the last layer, the activation function of sigmoid is applied for the last layer, and the update rule is Adam.
  The dimension of $\bm{p}$ is set to be sufficiently smaller than the number of demonstrations.
  The execution period of \equref{eq:rnnpb} is set to 5Hz.
}%
{%
  本研究で扱うネットワークであるrecurrent neural network with parametric bias (RNNPB)は, 以下のように式で表せる.
  基本のネットワーク構造は\cite{tani2002parametric}と同じである.
  \begin{align}
    (\bm{s}_{t+1}, \bm{u}_{t+1}) = \bm{h}(\bm{s}_{t}, \bm{u}_{t}, \bm{p}) \label{eq:rnnpb}
  \end{align}
  ここで, $t$は現在のタイムステップ, $\bm{s}$はセンサ状態, $\bm{u}$は制御入力, $\bm{p}$はparametric bias, $\bm{h}$はRNNPBを表す.
  $\bm{s}$や$\bm{u}$はロボットごとに異なり, 軸駆動型のロボットであれば$\bm{u}$は関節角度指令値$\bm{\theta}^{ref}$やトルク指令値$\bm{\tau}^{ref}$が一般的に使われる.
  腱駆動型のロボットによる模倣学習はこれまでに例がないが, 本研究では筋長指令値$\bm{l}^{ref}$を$\bm{u}$として用いている.
  $\bm{s}$は画像情報の他, 関節トルク$\bm{\tau}$や筋張力$\bm{f}$を入れても良い.
  本研究では, 画像情報を入れる際は一度AutoEncoder \cite{hinton2006reducing}を通して次元を圧縮し, $\bm{s}$に用いる.
  parametric bias $\bm{p}$は一つのdemonstration内では同一だが, 異なるdemonstration間では異なる学習可能な変数である.
  それぞれのセンサ値は全て得られたデータを使って正規化してからネットワークに入力している.

  本研究においてRNNPBは10層とし, 順に4層のfully-connected layer, 2層のLSTM layer \cite{hochreiter1997lstm}, 4層のfully-connected layerからなる.
  ユニット数については, \{$N_u+N_s+N_p$, 500, 300, 100, 100 (LSTMのunit数), 100 (LSTMのunit数), 100, 300, 500, $N_u+N_s$\}とした(なお, $N_{\{u, s, p\}}$は$\{\bm{u}, \bm{s}, \bm{p}\}$の次元数とする).
  activation functionはhyperbolic tangent, 更新則はAdam \cite{kingma2015adam}とした.
  画像を圧縮する際は, $96\times96$のRGB画像について, kernel sizeが3, strideが2の畳み込み層を5回適用し, 全結合層で順にユニット数256, 12まで次元を削減したあと, 同様に全結合層・逆畳み込み層によって画像を復元していく形を取っている.
  最終層以外についてはbatch normalization \cite{ioffe2015batchnorm}が適用され, activation functionは最終層以外についてはReLU \cite{nair2010relu}, 最終層はSigmoid, 更新則はAdam \cite{kingma2015adam}とした.
  $\bm{p}$の次元は, demonstrationの数よりも十分に小さいことを条件として, 実験ごとに適当に設定した.
  また, \equref{eq:rnnpb}の実行周期は5Hzとする.
}%

\subsection{Training} \label{subsec:training}
\switchlanguage%
{%
  We describe the training phase of RNNPB.
  For the simulation, a robot performs predetermined motions, and for the actual robot, a human teaches a task to the robot using motion capture, VR device, etc., and obtains the data of $\bm{s}$ and $\bm{u}$ at that time.
  For $k$-th demonstration of the task, we obtain the data $D_k=\{(\bm{s}_1, \bm{u}_{1}), (\bm{s}_2, \bm{u}_2), \cdots, (\bm{s}_{T_{k}}, \bm{u}_{T_{k}})\}$ ($1 \leq k \leq K$, where $K$ is the total number of demonstrations and $T_{k}$ is the number of time steps for the demonstration $k$).
  Then, we obtain the data $D_{train}=\{(D_1, \bm{p}_1), (D_2, \bm{p}_2), \cdots, (D_{K}, \bm{p}_K)\}$ for training.
  $\bm{p}_k$ is the parametric bias for the demonstration $k$, a vector that has the same value throughout one demonstration and has a different value in each demonstration.
  RNNPB is trained using this data $D_{train}$.
  In the training phase, we set $\bm{p}_{k}$ as trainable variables as well as the network weight $W$, i.e, we update not only $W$ as in the usual imitation learning but also $\bm{p}_{k}$ at the same time.
  Note that the loss function for training is the mean squared error between obtained and predicted network outputs, and all $\bm{p}_k$ are optimized with the initial value of $\bm{0}$.

  In the usual imitation learning, the average behavior among the obtained demonstrations is executed, but in this study, $\bm{p}_{k}$ reflects the motion style of the human for the demonstration $k$.
  For example, when manipulating an object, the position of the elbow and the speed of task execution are different for each person.
  Therefore, by controlling $\bm{p}$, it is possible to control motion style, though within the range of variation of human demonstrations.
  It should be noted that, although the actual motion style may change during the course of the demonstration even within one demonstration $k$, in this study, all motion styles in one demonstration are considered as one motion style.

  In the same way, $\bm{p}$ allows us to take into account the change in robot configurations such as muscle friction and muscle route due to the irreproducibility of initialization and deterioration over time \cite{kawaharazuka2020dynamics}, which is not the main topic of this study but is often observed in flexible bodies such as musculoskeletal humanoids.
  This enables the robot to continue moving correctly considering the sequential changes in the body.
  By acquiring data while changing the robot configuration, we can embed these changes into the parametric bias.
  At task execution, we can add constraints to make the network match the current robot configuration by changing $\bm{p}$.
}%
{%
  構築したRNNPBの学習方法について述べる.
  Simulationについては人間が予め決めた動作を行い, 実機についてはVR Device等を用いて人間がロボットに動作を教示し, その際の$\bm{s}$, $\bm{u}$のデータを取得していく.
  一回の試行$k$について, データ$D_k=\{(\bm{s}_1, \bm{u}_{1}), (\bm{s}_2, \bm{u}_2), \cdots, (\bm{s}_{T_{k}}, \bm{u}_{T_{k}})\}$を得る($1 \leq k \leq K$, $K$は全試行回数, $T_{k}$はその試行$k$に関する動作ステップ数とする).
  そして, 学習に用いるデータ$D_{train}=\{(D_1, \bm{p}_1), (D_2, \bm{p}_2), \cdots, (D_{K}, \bm{p}_K)\}$を得る.
  $\bm{p}_k$はその試行$k$に関するparametric biasであり, その一回の試行中については共通の値で, 異なる試行については別の値となる変数である.
  このデータ$D_{train}$を用いてRNNPBを学習させる.
  通常の模倣学習のようにネットワークの重み$W$のみを更新するのではなく, 同時に$\bm{p}_{k}$も更新していく.
  なお, 学習の際の損失関数はmean squared errorであり, 全$\bm{p}_k$は初期値を0として最適化される.

  通常の模倣学習では学習された中における平均的な動作が実行されるが, $\bm{p}_{k}$にはその試行における人間の動作スタイルが反映されることになる.
  例えば, 冗長性ゆえに手先で物体を操作するときは肘の角度は人それぞれであり, その動作の実行速度も人それぞれである.
  よって, $\bm{p}$を変化させることで, デモンストレーションのばらつきの範囲内ではあるが, 追加の制約を考慮することが可能となる.
  なお, 実際にはある試行$k$内でも途中で動作スタイルが変わることがあるが, 本研究ではそれらをまとめて一つのスタイルとしてしまっている.

  また同様に$\bm{p}$によって, 本研究の本題ではないが, 筋骨格ヒューマノイドのような柔軟身体で発生しやすい, initializationの再現性の無さによる状態の違いや経年変化等\cite{kawaharazuka2020dynamics}についても考慮することができる.
  これにより, ロボットの逐次的変化を考慮して正しく動き続けることが可能になる.
  ロボットのconfigurationを変化させながらデータを取得していくことで, それらの変化をparametric biasに埋め込むことができる.
  タスク実行時には, $\bm{p}$を変化させることで, ネットワークを現在のロボット状態に合致させる制約を追加することが可能となる.
}%

\subsection{Task Execution and Update of Parametric Bias} \label{subsec:update}
\switchlanguage%
{%
  The method of task execution is simple.
  All we have to do is to obtain the current sensor state $\bm{s}_{t}$ and control command $\bm{u}_{t}$, input them into \equref{eq:rnnpb} to obtain $\bm{u}_{t+1}$, and send it to the actual robot.

  The parametric bias is updated either offline or online at the same time as the task is executed.
  In this study, the difference in parametric bias due to the change in robot configuration as described in \secref{subsec:training} is denoted as (A), and the difference in parametric bias due to the variation of the motion style in human demonstrations is denoted as (B).
  For (A), since the state transition model ($(\bm{s}_{t}, \bm{u}_{t}) \rightarrow \bm{s}_{t+1}$) changes by the change in the robot configuration, $\bm{p}$ is updated to match the predictive models of \equref{eq:rnnpb} and the actual robot from the data of $(\bm{s}, \bm{u})$ when the robot actually moves.
  For (B), we need to consider additional constraints such as minimizing muscle tension or maximizing velocity.
  This means that (A) directly affects the task achievement, while (B) is a null space of the task achievement.
  Thus, once we execute the task with $\bm{p}=\bm{0}$ and the data $D=\{(\bm{s}^{data}_1, \bm{u}^{data}_{1}), (\bm{s}^{data}_2, \bm{u}^{data}_2), \cdots, (\bm{s}^{data }_{T}, \bm{u}^{data}_{T})\}$ is obtained ($T$ is the number of time steps), we update $\bm{p}$ by defining the loss functions for A and B as follows,
  \begin{align}
    L = ||\bm{s}^{data}_{2:T}-\bm{s}^{pred}_{2:T}||_2 + \alpha{L}_{constraint}(\bm{x}^{pred'}_{2:T}) \label{eq:loss}
  \end{align}
  where $||\cdot||_2$ is L2 norm, $\bm{x}$ is $\begin{pmatrix}\bm{s}^{T}&\bm{u}^{T}\end{pmatrix}^{T}$, $\bm{x}_{a:b}$ is the sequence of $\bm{x}$ between $[a, b]$ time steps, $\bm{s}^{pred}_{2:T}$ is the sequence of $\bm{s}$ obtained by passing $\bm{x}^{data}_{1:T-1}$ through $\bm{h}$ in order, $\alpha$ is the weight, and $L_{constraint}$ is the loss function for the additional constraint in (B).
  Also, $\bm{x}^{pred'}_{2:T}$ is the value of $\bm{x}$ obtained by repeating the autoregressive process, passing $\bm{x}^{data}_{1}$ through $\bm{h}$, and inputting the output into $\bm{h}$ recursively.
  The first term on the right-hand side of \equref{eq:loss} is the loss for (A), where the output $\bm{u}_{t+1}$ from \equref{eq:loss} is ignored and $\bm{p}$ is updated to minimize the error of the predictive model of $\bm{s}$.
  The second term on the right-hand side of \equref{eq:loss} is the loss for (B), where since $\bm{u}_{t+1}$ changes as $\bm{p}$ changes, only the initial value of data $D$ is used and the calculation must proceed autoregressively.
  If we take the loss only for (B), we do not need to execute the task, and only the current state $(\bm{s}^{data}_1, \bm{u}^{data}_{1})$ is needed to be given.
  There are various possible definitions for $L_{constraint}$, examples of which are shown below,
  \begin{align}
    L^{tension}_{constraint} &= ||\bm{f}^{pred}_{2:T}||_{2} \label{eq:tension}\\
    L^{lvelocity}_{constraint} &= ||\bm{l}^{pred}_{3:T}-\bm{l}^{pred}_{2:T-1}||_{2} \label{eq:lvelocity}\\
    L^{jvelocity}_{constraint} &= ||\bm{\theta}^{pred}_{3:T}-\bm{\theta}^{pred}_{2:T-1}||_{2} \label{eq:jvelocity}
  \end{align}
  where $\bm{f}^{pred}$ is the value of $\bm{x}^{pred}$ when $\bm{x}^{pred}$ contains the muscle tension $\bm{f}$, $\bm{l}^{pred}$ is the value of $\bm{x}^{pred}$ when $\bm{x}^{pred}$ contains the muscle length $\bm{l}$, and $\bm{\theta}^{pred}$ is the value of $\bm{x}^{pred}$ when $\bm{x}^{pred}$ contains the joint angle $\bm{\theta}$.
  If we want to minimize the muscle tension, \equref{eq:tension} is used with positive $\alpha$, and if we want to increase the muscle tension and maintain the joint stiffness, \equref{eq:tension} is used with negative $\alpha$, which is also true for the velocity.
  We can also constrain $\bm{p}$ not to converge to an undesired value by adding the term $||\bm{p}||_{2}$ as a constraint to \equref{eq:loss}.
  Since $\bm{p}_{k}$ basically concentrates around $\bm{0}$, we can add a minimization constraint of $\bm{p}$ to prevent it from deviating too far from $\bm{0}$.
  When updating $\bm{p}$, the network weight $W$ is fixed and only $\bm{p}$ is updated with Momentum SGD \cite{qian1999momentum}.
  When updating the parametric bias offline, we execute the task with $\bm{p}=\bm{0}$ once and update $\bm{p}$ using the data $D$ obtained in the execution.
  When updating parametric bias online, $\bm{p}$ is updated using the accumulated data $D$ after the number of data $N^{online}$ exceeds the threshold $N^{online}_{thre}$.
  The data exceeding $N^{online}_{max}$ are discarded from the oldest ones.

  In this study, as shown in \equref{eq:tension}--\equref{eq:jvelocity}, we are only dealing with minimization and maximization of velocity and force.
  Theoretically, it is possible to give a certain desired value and update $\bm{p}$ so that the velocity and force become close to it.
  However, when opening a box, for example, the speed is not always constant but largely varies even in a single demonstration.
  Even if a certain desired value is given, it will not be realized, and only the overall speed becomes close to the desired value.
  Therefore, in this study, the values are manipulated by minimization and maximization.
  Note that the the magnitude of the velocity and force can be roughly controlled by scaling $\alpha$ in \equref{eq:loss} accordingly.

  In this study, $\alpha$ and $L_{constraint}$ are different depending on the experiment, and $N^{online}_{thre}=10$ and $N^{online}_{max}=20$ are used.
}%
{%
  タスクの実行方法はシンプルである.
  現在のセンサ状態$\bm{s}_{t}$と制御入力$\bm{u}_{t}$を取得し, \equref{eq:rnnpb}を順伝播することで$\bm{u}_{t+1}$を得て, これを実機に送ることを繰り返すのみである.

  このタスク実行と同時にオフラインまたはオンラインでparametric biasの更新も行う.
  本研究では, \secref{subsec:training}で述べた, ロボットのconfiguration変化によるparametric biasの違いを(A), 人間のデモンストレーションにおける動作スタイルのばらつきによるparametric biasの違いを(B)とする.
  (A)については, ロボットの状態変化によって状態遷移モデル($(\bm{s}_{t}, \bm{u}_{t}) \rightarrow \bm{s}_{t+1}$)が変化するため, 実際に動作したときの$(\bm{s}, \bm{u})$と合致するように, $\bm{p}$を更新する.
  (B)については, 筋張力最小化や速度最大化等, 追加の制約を考える必要がある.
  これは, (A)はtask達成に直接影響するのに対して, (B)はtask達成に関するnull spaceであるということである.
  よって, 一度$\bm{p}=\bm{0}$でタスクを実行してデータ$D=\{(\bm{s}^{data}_1, \bm{u}^{data}_{1}), (\bm{s}^{data}_2, \bm{u}^{data}_2), \cdots, (\bm{s}^{data}_{T}, \bm{u}^{data}_{T})\}$が得られた時に, AとBについて以下のように損失関数を定義することで$\bm{p}$を更新していく.
  \begin{align}
    L = ||\bm{s}^{data}_{2:T}-\bm{s}^{pred}_{2:T}||_2 + \alpha{L}_{constraint}(\bm{x}^{pred'}_{2:T}) \label{eq:loss}
  \end{align}
  ここで, $||\cdot||_2$はL2ノルム, $\bm{x}$は$\begin{pmatrix}\bm{s}^{T}&\bm{u}^{T}\end{pmatrix}^{T}$, $\bm{x}_{a:b}$は$[a, b]$の間の$\bm{x}$を並べたもの, $\bm{s}^{pred}_{2:T}$は$\bm{x}^{data}_{1:T-1}$を順に$\bm{h}$に通して得られた$\bm{s}$の値, $\alpha$は重みの係数, $L_{constraint}$は(B)の追加制約に関する損失関数を表す.
 また, $\bm{x}^{pred'}_{2:T}$は$\bm{x}^{data}_{1}$から始め, $\bm{h}$に通し, その出力を$\bm{h}$に入力するという自己回帰を繰り返した際に得られた$\bm{x}$の値である.
  \equref{eq:loss}の右辺第一項は(A)に関する損失であり, \equref{eq:rnnpb}の出力$\bm{u}_{t+1}$を無視し, $\bm{s}$の予測モデル誤差を最小化するように$\bm{p}$を更新する.
  \equref{eq:loss}の右辺第二項は(B)に関する損失であり, $\bm{p}$が変化することによって入力の$\bm{u}$が変化するため, データ$D$は初期値のみしか用いず, 自己回帰によって計算を進める必要がある.
  もし(B)のみに関して損失を取るのであれば, 一度動作を試行する必要はなく, 現在状態$(\bm{s}^{data}_1, \bm{u}^{data}_{1})$のみ与えても良い.
  $L_{constraint}$には様々な定義が可能だが, その一例を以下に示す.
  \begin{align}
    L^{tension}_{constraint} &= ||\bm{f}^{pred}_{2:T}||_{2} \label{eq:tension}\\
    L^{lvelocity}_{constraint} &= ||\bm{l}^{pred}_{3:T}-\bm{l}^{pred}_{2:T-1}||_{2} \label{eq:lvelocity}\\
    L^{jvelocity}_{constraint} &= ||\bm{\theta}^{pred}_{3:T}-\bm{\theta}^{pred}_{2:T-1}||_{2} \label{eq:jvelocity}
  \end{align}
  ここで, $\bm{f}^{pred}$は$\bm{x}^{pred}$が筋張力$\bm{f}$を含む場合のその値, $\bm{\theta}^{pred}$は$\bm{x}^{pred}$が関節角度$\bm{\theta}$を含む場合のその値である.
  筋張力を最小化したい場合は$\alpha$を正として\equref{eq:tension}を用い, 筋張力を高めにして剛性を保ちたい場合は$\alpha$を負とすれば良く, これは速度についても同様である.
  この他にも, \equref{eq:loss}に制約として$||\bm{p}||_{2}$の項を加えることで, $\bm{p}$がundesired valueに収束しないように制約することもできる.
  $\bm{p}_{k}$は基本的に$\bm{0}$周辺に集まり, $\bm{p}=\bm{0}$のときが最も平均的な動作になるので, そこから大きく外れないように最小化制約を追加する.
  実際に$\bm{p}$を更新する際は, ネットワークの重み$W$を固定して$\bm{p}$のみMomentum SGD \cite{qian1999momentum}で更新する.
  オフラインでparametric biasを更新する際は, 一度$\bm{p}=\bm{0}$の状態で動作を行い, その際に得られたデータ$D$を用いて$\bm{p}$を更新する.
  オンラインでparametric biasを更新する際は, データ数$N^{online}$が閾値$N^{online}_{thre}$を超えてから, 蓄積したデータ$D$を用いて$\bm{p}$を更新する.
  なお, $N^{online}_{max}$を超えたデータは古いものから破棄していく.

  本研究では\equref{eq:tension}--\equref{eq:jvelocity}にあるように, 速度や力の最小化・最大化のみを扱っているが, 実際にはある指令値を与え, これに速度や力が近くなるように$\bm{p}$を更新することも可能である.
  しかし, 例えば箱を開けるときはその一つの動作でも常に速度は一定ではなく様々に変化するため, ある指令値を与えても, それが実現できるわけではなく, 全体的に見たときの速度が大体指令値に近い, という程度にしかならない.
  ゆえに, 本研究では最小化・最大化によって値を操作し, そのときの係数$\alpha$の大きさによって, 大まかに速度に関するdesired valueを与えるという形にしている.

  本研究では, $\alpha$や$L_{constraint}$は実験によって異なり, $N^{online}_{thre}=10$, $N^{online}_{max}=20$とした.
}%

\begin{figure}[t]
  \centering
  \includegraphics[width=0.9\columnwidth]{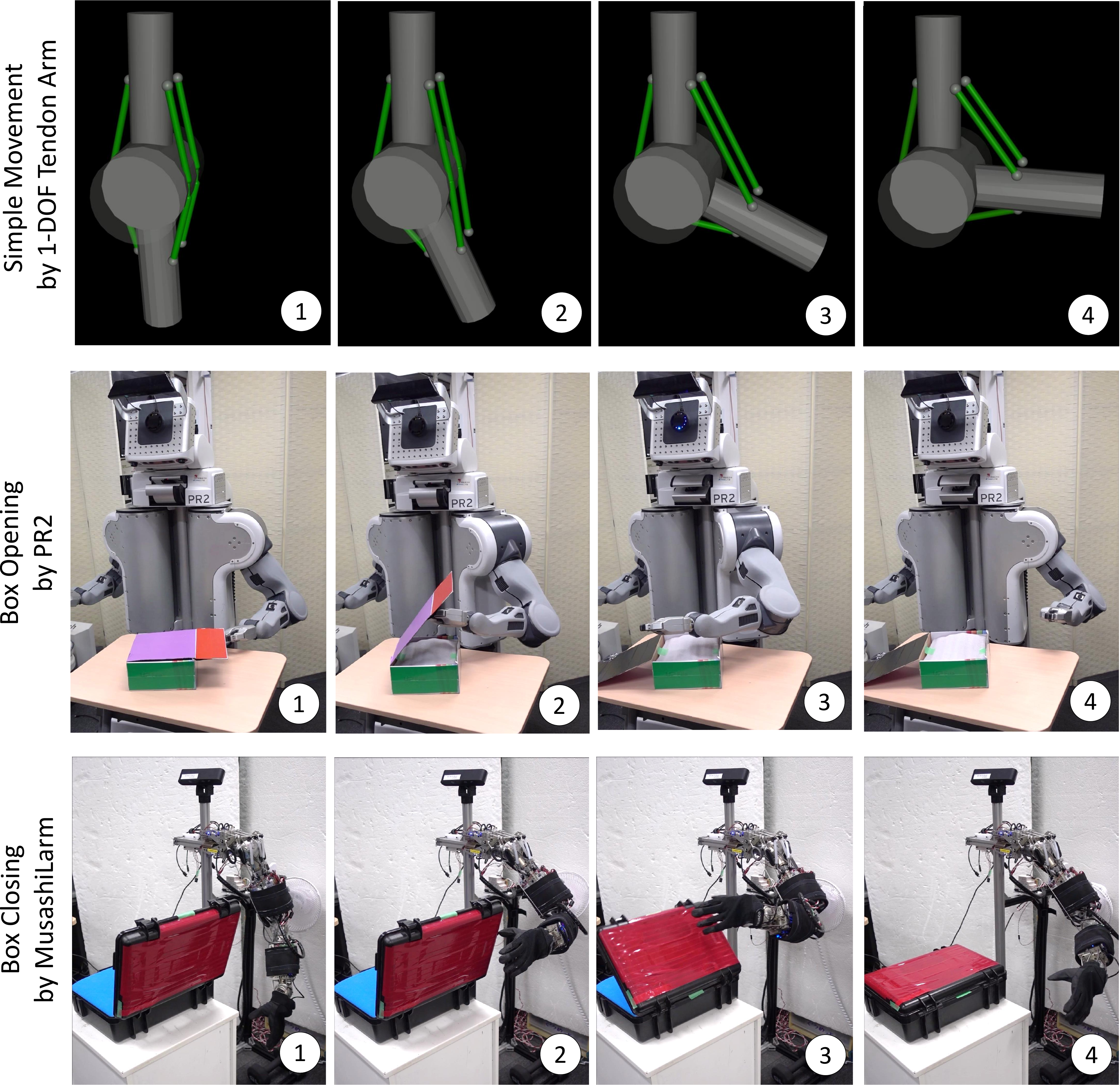}
  \vspace{-1.0ex}
  \caption{Experimental setup: simple movement by 1-DOF tendon arm, box opening by PR2, and box closing by MusashiLarm.}
  \label{figure:experimental-setup}
  \vspace{-3.0ex}
\end{figure}

\section{Experiments} \label{sec:experiments}
\subsection{Experimental Setup} \label{subsec:experimental-setup}
\switchlanguage%
{%
  \figref{figure:experimental-setup} shows each of the experimental tasks to be executed.

  First, we simulate a simple tendon arm with one degree of freedom and three muscles to validate the effectiveness of the proposed method.
  This simulates nonlinear elastic elements and the muscle friction which make the dynamics model complex.
  The task is a simple movement to move the joint angle from 0 to -90 degrees.
  The demonstration behavior is a feedback control that repeatedly updates the target joint angle and target muscle tension in the form of $\bm{\theta}^{ref}_{t+1} \gets \bm{\theta}^{ref}_{t} + \beta(\bm{\theta}^{task}-\bm{\theta}_{t})$, $\bm{f}^{ref}_{t+1} \gets \bm{f}^{ref}_{t} + \beta(\bm{f}^{style}-\bm{f}^{ref}_{t})$ and sends the target muscle length to the robot in the form of $\bm{l}^{ref}=\bm{h}_{bodyimage}(\bm{\theta}^{ref}, \bm{f}^{ref})$ ($\bm{\theta}^{task} = -90$ [deg] represents the joint angle to be achieved, $\bm{\theta}_{t}$ represents the measured current joint angle, $\bm{f}^{style}$ represents the muscle tension to be achieved as the motion style, and $\beta$ is a coefficient that represents the rate of feedback).
  Here, $\bm{h}_{bodyimage}$ is the conversion function from the target joint angle $\bm{\theta}^{ref}$ and target muscle tension $\bm{f}^{ref}$ to the target muscle length $\bm{l}^{ref}$ learned by \cite{kawaharazuka2019longtime}.
  Note that since this model only considers the static sensor relationship, sending $\bm{l}^{ref}=\bm{h}_{bodyimage}(\bm{\theta}^{task}, \bm{f}^{ref})$ does not directly realize $\bm{\theta}^{task}$ and so this feedback control is used.
  The velocity and muscle tension can be adjusted by arbitrarily selecting $\beta$ and $\bm{f}^{style}$, and the method is verified by varying these parameters ($\bm{f}^{ref}$ can be set independently for each muscle, and the muscle tension should be set to compensate for gravity. However, in this study, the target muscle tension for all muscles is set to the same value for simplicity. The changes in $\beta$ and $\bm{f}^{style}$ correspond to (B) in \secref{subsec:update}).
  In addition, by changing the joint radius $r$ of the 1-DOF joint, i.e. the moment arm of muscles, it is possible to produce the change in robot configuration, which is (A) of \secref{subsec:update} and is also used for verification.
  Note that in RNNPB, $\bm{s} = \begin{pmatrix} \bm{\theta}^{T} & \bm{f}^{T} \end{pmatrix}^{T}$, $\bm{u}=\bm{l}^{ref}$, and $\bm{p}$ is two-dimensional (in this experiment, the dimension of $\bm{p}$ is set to match the dimension of the motion style of the demonstration for the sake of clarity of plots).

  Next, we conduct experiments on an axis-driven dual-arm robot PR2.
  The task is to open a box on a table using the left arm with 7-DOF.
  The right arm is controlled only with the gravity compensation torque, and the left arm is controlled to send the joint angle symmetric to the right arm.
  When a human moves the right arm of PR2, the left arm of PR2 demonstrates the task, and imitation learning is performed.
  In this process, we rotate the box in various directions to obtain data.
  Although we used demonstration data from a single person who performed the task many times and did not add explicit perturbation, $\bm{p}$ varied because the behavior of the demonstrator changed throughout the demonstrations.
  Note that in RNNPB, $\bm{s} = \bm{z}$, $\bm{u} = \bm{\theta}^{ref}$ ($\bm{z}$ is the image compressed by AutoEncoder), and $\bm{p}$ is three-dimensional.

  Finally, we conduct experiments on the musculoskeletal humanoid MusashiLarm \cite{kawaharazuka2019musashi}.
  The task is to close a box on a table using the left arm with 7-DOF and 18 muscles.
  We obtain the hand coordinates from HTC VIVE controller, which is operated by a human, and solve inverse kinematics to obtain the target joint angle $\bm{\theta}^{ref}$, send the target muscle length by $\bm{l}^{ref}=\bm{h}_{bodyimage}(\bm{\theta}^{ref}, \bm{f}^{ref})$, and perform the demonstration (here, we assume that $\bm{f}^{ref}=10$ [N] is constant).
  In this process, we rotate the box in various directions to obtain data.
  Note that in RNNPB, $\bm{s} = \begin{pmatrix}\bm{z}^{T} & \bm{f}^{T}\end{pmatrix}^{T}$, $\bm{u}=\bm{l}^{ref}$, and $\bm{p}$ is three-dimensional.
}%
{%
  \figref{figure:experimental-setup}にそれぞれの実験のセットアップを示す.

  まず, 1自由度3筋の簡易的な筋骨格ロボットのシミュレーションにより, 基本的な性能を示す.
  本シミュレーションは筋に非線形弾性ばねの要素と摩擦の影響を入れ, モデル化を困難にしている.
  タスクは関節角度を0から-90度まで動かすという簡単な動きである.
  デモンストレーションの動作は, $\bm{\theta}^{ref}_{t+1} \gets \bm{\theta}^{ref}_{t} + \beta(\bm{\theta}^{task}-\bm{\theta}_{t})$, $\bm{f}^{ref}_{t+1} \gets \bm{f}^{ref}_{t} + \beta(\bm{f}^{style}-\bm{f}^{ref}_{t})$という形で指令関節角度と指令筋張力を更新し, $\bm{l}^{ref}=\bm{h}_{bodyimage}(\bm{\theta}^{ref}, \bm{f}^{ref})$の形で指令筋長を送ることを繰り返すフィードバック制御である($\bm{\theta}^{task} = -90$ [deg]は実現したい関節角度, $\bm{\theta}_{t}$は測定された現在関節角度, $\bm{f}^{style}$は教示のスタイルとしての筋張力指令, $\beta$はフィードバックの割合を表す係数を表す).
  ここで, $\bm{h}_{bodyimage}$は\cite{kawaharazuka2019longtime}によって学習される指令関節角度・指令筋張力$\bm{f}^{ref}$から筋長への変換関数である.
  なお, このモデルは静的な関係しか考慮していないため, 直接$\bm{l}^{ref}=\bm{h}_{bodyimage}(\bm{\theta}^{task}, \bm{f}^{ref})$を送っても$\bm{\theta}^{task}$を実現できないため, このようなフィードバック制御を行っている.
  $\beta$と$\bm{f}^{style}$を任意に選ぶことで速度や筋張力を調整でき, これを変化させて本手法の検証を行う($\bm{f}^{ref}$はそれぞれの筋について独立に設定でき, 重力補償可能な筋張力を設定すべきであるが, 本研究では簡単のため全筋に対する指令筋張力を同じ値としている. これらは, \secref{subsec:update}における(B)に相当する).
  また, 1自由度関節の半径$r$を変化させることで, \secref{subsec:update}の(A)のrobot configurationの変化を作り出すことが可能であり, これも検証に用いる.
  なお, RNNPBにおいて, $\bm{s} = \begin{pmatrix} \bm{\theta}^{T} & \bm{f}^{T} \end{pmatrix}^{T}$, $\bm{u}=\bm{l}^{ref}$であり, $\bm{p}$は2次元とした(本実験では, $\bm{p}$のプロットをわかりやすくするため, demonstrationのmotion styleの次元と一致させている).

  次に, 台車型双腕型ロボットPR2について実験を行う.
  タスクは, 7自由度の左腕を使って, 机の上の箱を開ける動作である.
  右手を重力補償トルクのみをかけた状態にし, 右手と対称な関節角度を左手に送るような制御を行う.
  人間がPR2の右手を動かすことでPR2の左手によりデモンストレーションを行い, 模倣学習を行う.
  その際, 箱を様々な方向に回転させてデータを取る.
  なお, RNNPBにおいて, $\bm{s} = \bm{z}$, $\bm{u} = \bm{\theta}^{ref}$であり($\bm{z}$は画像をAutoEncoderにより圧縮した値とする), $\bm{p}$は3次元とした.

  最後に, 筋骨格ヒューマノイドMusashiLarm \cite{kawaharazuka2019musashi}について実験を行う.
  タスクは, 7自由度18筋の左腕を使って, 机の上の箱を閉じる動作である.
  人間が操作するHTC Viveのcontrollerから手先座標を取得し, そこに対して逆運動学を解いて指令関節角度$\bm{\theta}^{ref}$を求め, $\bm{l}^{ref}=\bm{h}_{bodyimage}(\bm{\theta}^{ref}, \bm{f}^{ref})$により指令筋長を送ってデモンストレーションを実行する(ここでは$\bm{f}^{ref}=10$ [N]で一定としている).
  その際, 箱を様々な方向に回転させてデータを取る.
  なお, RNNPBにおいて, $\bm{s} = \begin{pmatrix} \bm{z}^{T} & \bm{f}^{T} \end{pmatrix}^{T}$, $\bm{u}=\bm{l}^{ref}$であり, $\bm{p}$は3次元とした.
}%

\begin{figure}[t]
  \centering
  \includegraphics[width=0.9\columnwidth]{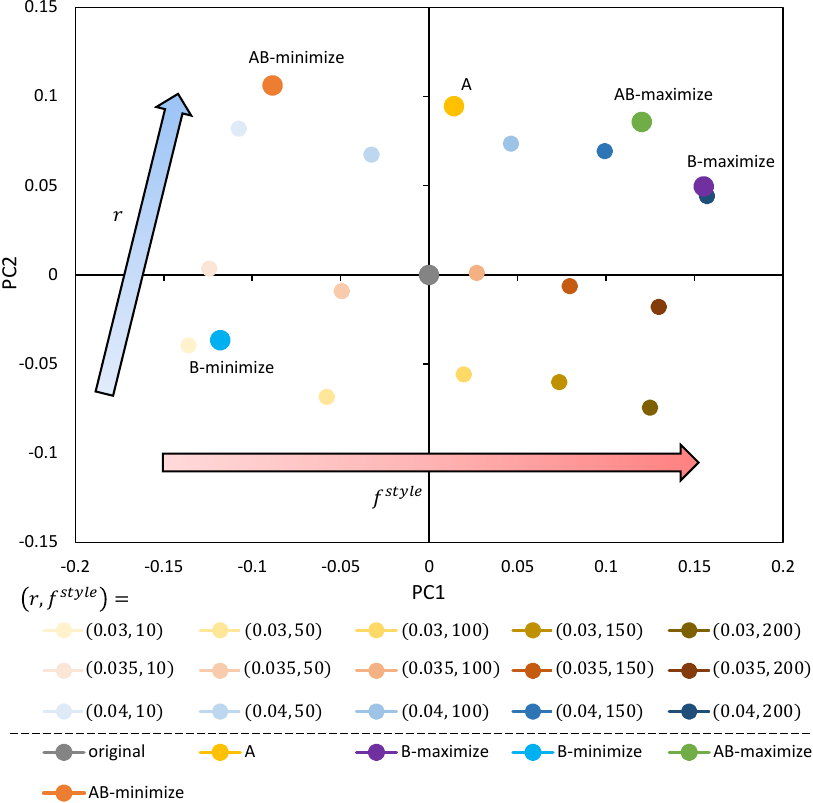}
  \vspace{-1.0ex}
  \caption{Simple movement experiment by simulated 1-DOF tendon arm when changing $\bm{r}$ and $\bm{f}^{style}$: parametric biases trained in data collection phase and those updated from additional constraints regarding ``A'', ``B-maximize'', ``B-minimize'', ``AB-maximize'', and ``AB-minimize''.}
  \label{figure:1dofsim-exp1-pb}
  \vspace{-1.0ex}
\end{figure}

\begin{figure}[t]
  \centering
  \includegraphics[width=1.0\columnwidth]{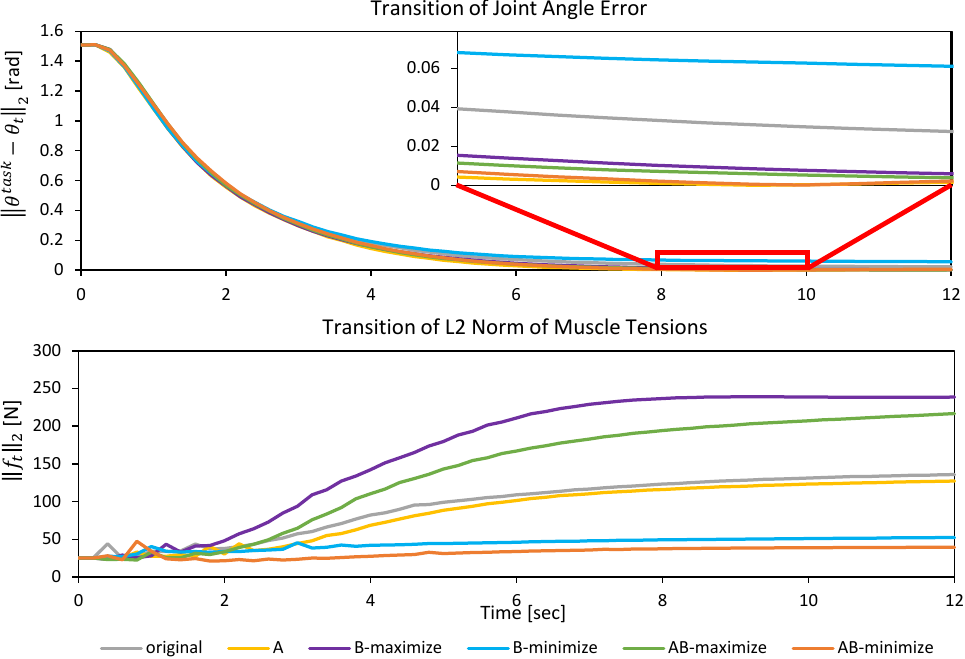}
  \vspace{-3.0ex}
  \caption{Evaluation of simple movement experiment by simulated 1-DOF tendon arm when changing $\bm{r}$ and $\bm{f}^{style}$: transition of $||\bm{\theta}^{task}-\bm{\theta}_{t}||_{2}$ and $||\bm{f}_{t}||_{2}$.}
  \label{figure:1dofsim-exp1-eval}
  \vspace{-3.0ex}
\end{figure}

\begin{figure}[t]
  \centering
  \includegraphics[width=0.9\columnwidth]{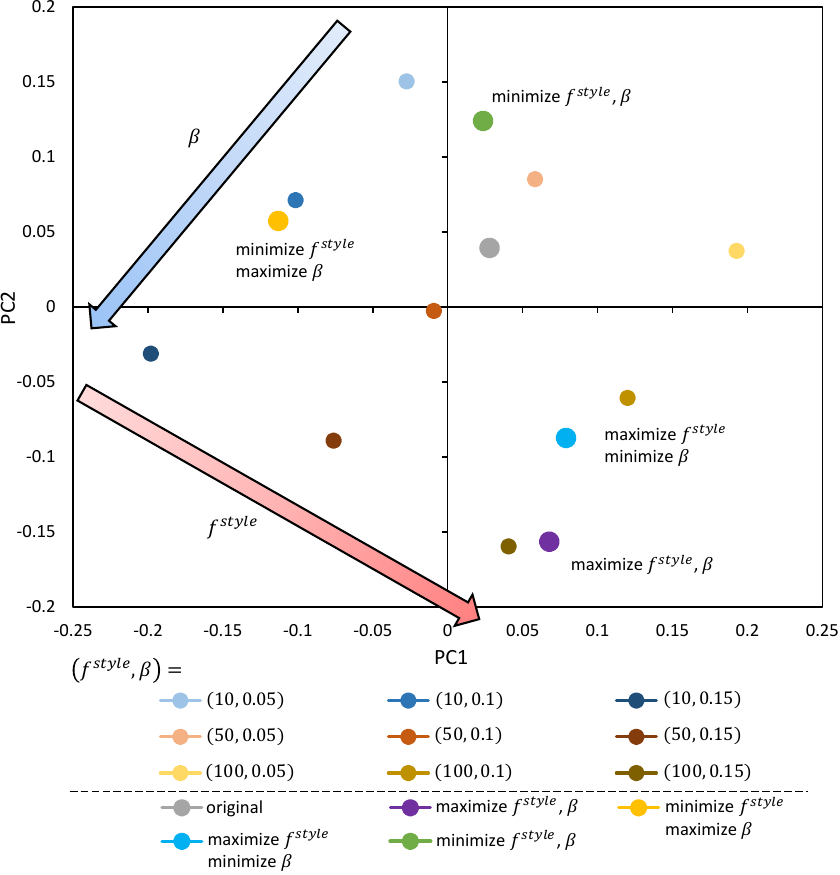}
  \vspace{-1.0ex}
  \caption{Simple movement experiment by simulated 1-DOF tendon arm when changing $\bm{f}^{style}$ and $\beta$: parametric biases trained in data collection phase and those updated from additional constraints.}
  \label{figure:1dofsim-exp2-pb}
  \vspace{-1.0ex}
\end{figure}

\begin{figure}[t]
  \centering
  \includegraphics[width=1.0\columnwidth]{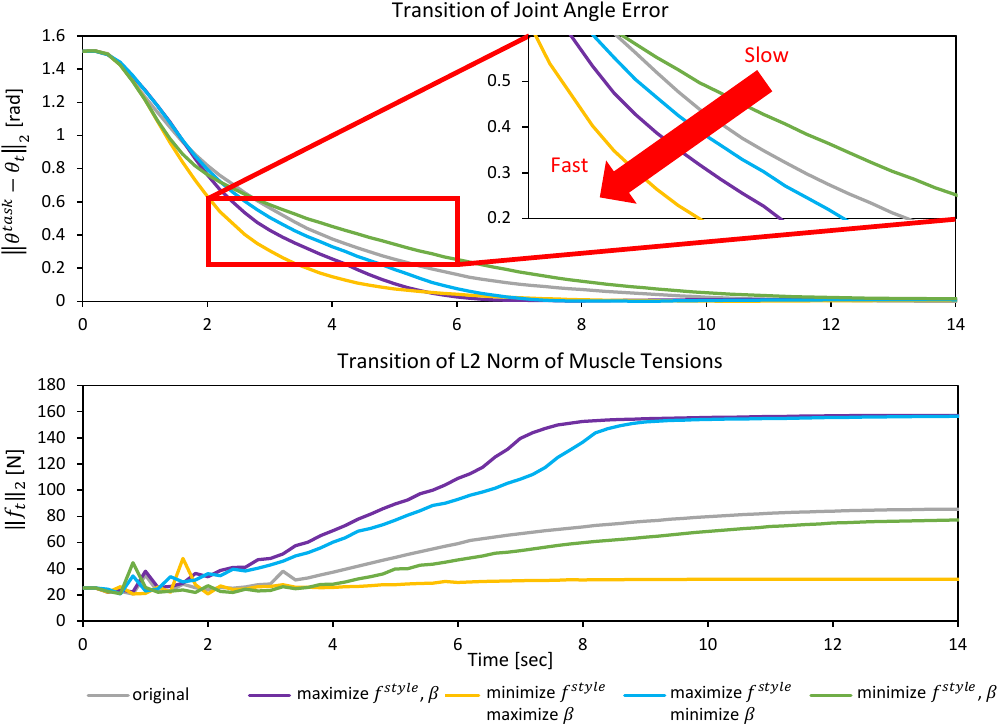}
  \vspace{-3.0ex}
  \caption{Evaluation of simple movement experiment by simulated 1-DOF tendon arm when changing $\bm{f}^{style}$ and $\beta$: transition of $||\bm{\theta}^{task}-\bm{\theta}_{t}||_{2}$ and $||\bm{f}_{t}||_{2}$.}
  \label{figure:1dofsim-exp2-eval}
  \vspace{-3.0ex}
\end{figure}

\begin{figure}[t]
  \centering
  \includegraphics[width=1.0\columnwidth]{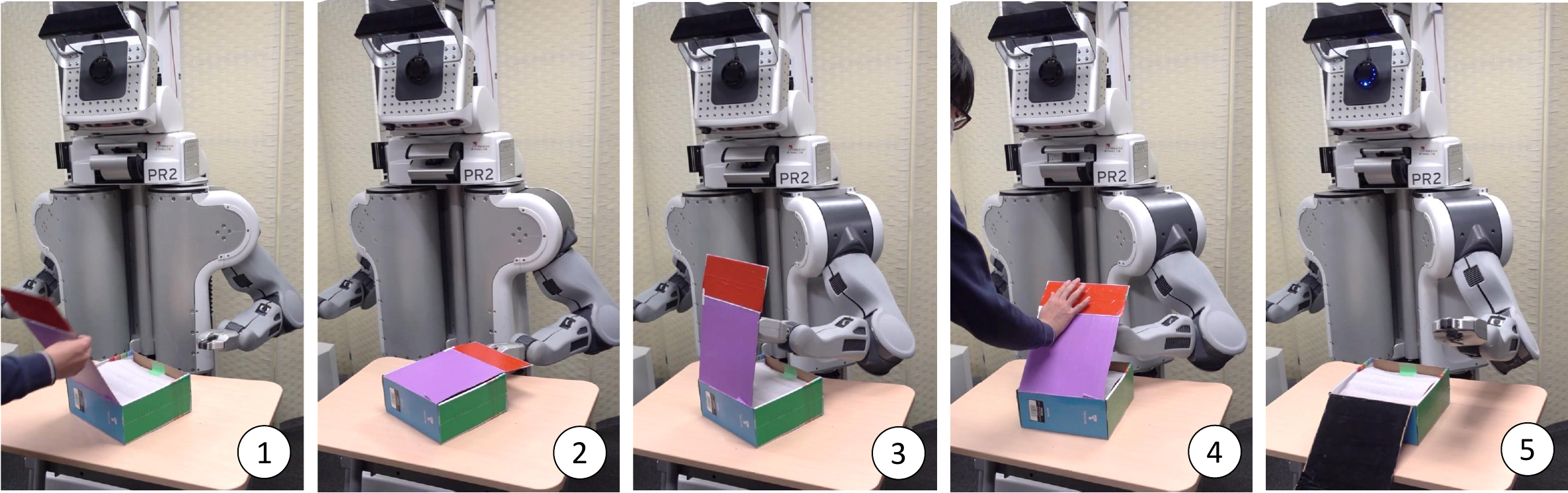}
  \vspace{-3.0ex}
  \caption{The trained behavior of box opening by PR2.}
  \label{figure:pr2-exp-behavior}
  \vspace{-1.0ex}
\end{figure}

\begin{figure}[t]
  \centering
  \includegraphics[width=0.7\columnwidth]{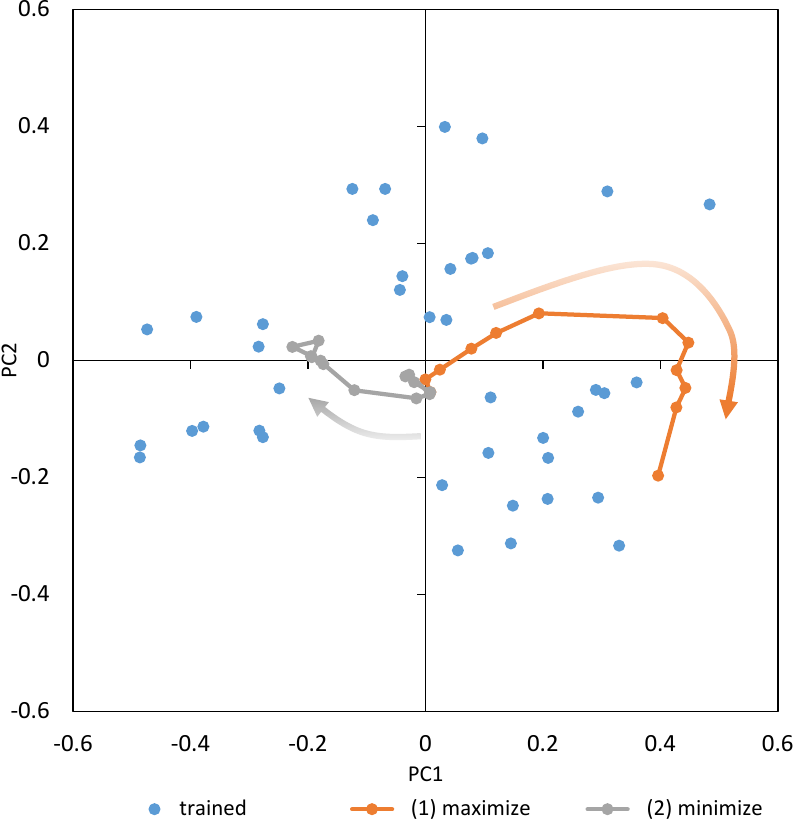}
  \vspace{-1.0ex}
  \caption{Box opening experiment by PR2: parametric biases trained in data collection phase and their trajectories updated from additional constraints regarding (1) and (2).}
  \label{figure:pr2-exp-pb}
  \vspace{-3.0ex}
\end{figure}

\begin{figure}[t]
  \centering
  \includegraphics[width=1.0\columnwidth]{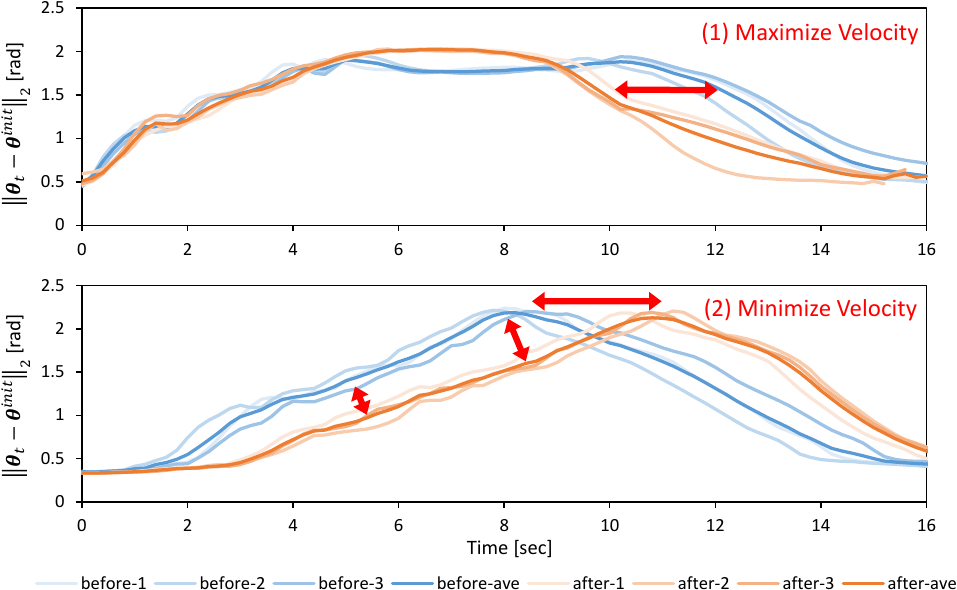}
  \vspace{-3.0ex}
  \caption{Evaluation of box opening experiment by PR2: transition of $||\bm{\theta}_{t}-\bm{\theta}^{init}||_{2}$ when using $\bm{p}$ before and after online update regarding (1) and (2).}
  \label{figure:pr2-exp-eval}
  \vspace{-3.0ex}
\end{figure}

\subsection{Simple Movement by Simulated 1-DOF Tendon Arm} \label{subsec:simulator-exp}
\switchlanguage%
{%
  First, experiments were conducted by varying the radius of the 1-DOF joint $\bm{r}$ and the muscle tension $\bm{f}^{style}$.
  For each of the 15 combinations of $r=\{0.03, 0.035, 0.04\}$ [m] and $\bm{f}^{style}=\{10, 50, 100, 150, 200\}$ [N], about 60 steps of data were obtained and used to train RNNPB.
  The two-dimensional representation of $\bm{p}_{k}$ obtained during training is shown in \figref{figure:1dofsim-exp1-pb} through principle component analysis (PCA).
  We can see that $\bm{p}_{k}$ is neatly aligned along the size of $r$ and $\bm{f}^{style}$.
  In \figref{figure:1dofsim-exp1-pb}, we also plotted parametric biases which are updated from additional constraints by applying \equref{eq:loss} offline denoted as, e.g. ``B-minimize'' or ``AB-maximize''.
  We first executed the task with $\bm{p}=\bm{0}$ and $r = 0.04$, and then we updated $\bm{p}$ by applying \equref{eq:loss} with different constraints on (B).
  Here, the change in $r$ corresponds to (A) in \secref{subsec:update}, and the change in $\bm{f}^{style}$ corresponds to (B).
  We refer to the combinations of whether to include the matching constraint on (A), and whether to maximize, minimize, or not include the constraint on (B): ``A'', ``B-maximize'', ``B-minimize'', ``AB-maximize'', and ``AB-minimize''.
  When updating $\bm{p}$, $\alpha=0.1$ (the sign is reversed for maximization), the learning rate is 0.01, the epoch is 30, and \equref{eq:tension} is used as the loss function for (B).
  Also, ``original'' expresses $\bm{p}=0$ (note that $\bm{p}=\bm{0}$ is not necessarily at the origin of the graph, since it is transformed by PCA).
  All $\bm{p}$ considering (A) have values close to the trained $\bm{p}_{k}$ at $r=0.04$, indicating that the current value of $\bm{r}$ is correctly recognized.
  Also, considering (B), we can see that $\bm{p}$ has moved to the vicinity of the trained $\bm{p}_{k}$ at $\bm{f}^{style}=200$ [N] for maximization and to the vicinity of the trained $\bm{p}_{k}$ at $\bm{f}^{style}=10$ [N] for minimization.
  $||\bm{\theta}^{task}-\bm{\theta}_{t}||_{2}$ and $||\bm{f}_{t}||_{2}$ during the task using $\bm{p}$ obtained here are shown in \figref{figure:1dofsim-exp1-eval}.
  Note that while $\bm{p}$ is modified continuously during online update experiments, $\bm{p}$ before and after the online update are fixed during evaluation experiments such as \figref{figure:1dofsim-exp1-eval}.
  The evaluation experiment is conducted without online update to obtain consistent graphs.
  For $\bm{p}$ of ``original'' and ``B-minimize'' far from the trained $\bm{p}_{k}$ at $r=0.04$, $\bm{\theta}_{t}$ does not reach $\bm{\theta}^{task}$, indicating that there is an error in achieving the task.
  Also, $\bm{f}_{t}$ is small for $\bm{p}$ that minimizes (B), $\bm{f}_{t}$ is large for $\bm{p}$ that maximizes (B), and $\bm{f}_{t}$ is somewhere in between for $\bm{p}$ with no constraint on (B).
  For $\bm{p}$ of ``B-maximize'' without the constraint on (A), we obtained $\bm{p}$ near the trained $\bm{p}_{k}$ at $r=0.04$ by coincidence, and the behavior is similar to that of ``AB-maximize''.

  Next, the experiment was conducted by varying the muscle tension $\bm{f}^{style}$ and the coefficient $\beta$ representing the motion speed ($r=0.04$ was fixed).
  For each of the nine combinations of $\bm{f}^{style}=\{10, 50, 100\}$ [N] and $\beta=\{0.05, 0.10, 0.15\}$, about 60 steps of data were obtained and used to train RNNPB.
  The two-dimensional representation of $\bm{p}_{k}$ obtained during training is shown in \figref{figure:1dofsim-exp1-pb} through PCA.
  We can see that $\bm{p}_{k}$ is neatly aligned along the size of $\bm{f}^{style}$ and $\beta$.
  The values of $\bm{p}$ updated offline from additional constraints as in the previous experiment are also shown in \figref{figure:1dofsim-exp2-pb}.
  Here, we experimented with four combinations of $\bm{f}^{style}$ and $\beta$, using the loss functions of \equref{eq:tension} and \equref{eq:lvelocity}, to maximize or minimize each of them.
  When updating $\bm{p}$, the weights are $(\alpha_1, \alpha_2)=(0.02, 0.3)$ (the weight for $\bm{f}^{style}$ is $\alpha_{1}$ and for $\beta$ is $\alpha_{2}$), the learning rate is 0.01, and the epoch is 30.
  In the following experiments, we basically consider only the constraints on (B), but we update $\bm{p}$ including the matching constraint on (A) as in \equref{eq:loss}.
  In the axis of $\bm{f}^{style}$ in \figref{figure:1dofsim-exp2-pb}, $\bm{p}$ is updated relatively as intended by minimization and maximization.
  In contrast, in the axis of $\beta$, not all $\bm{p}$ were updated in the intended direction.
  Although the desired results can be obtained by adjusting $\alpha_1$ and $\alpha_2$ for each optimization, it was found that careful adjustment of the ratio of weights is necessary in order to consider more than two constraints on (B).
  $||\bm{\theta}^{task}-\bm{\theta}_{t}||_{2}$ and $||\bm{f}_{t}||_{2}$ for the task with the obtained $\bm{p}$ are shown in \figref{figure:1dofsim-exp2-eval}.
  As with the placement of $\bm{p}$ on the axis of $\bm{f}^{style}$ in \figref{figure:1dofsim-exp2-pb}, $||\bm{f}_{t}||_{2}$ can be controlled as intended by minimization and maximization in most cases.
  Only when both $\bm{f}^{style}$ and $\beta$ were minimized, $\bm{f}^{style}$ was not fully minimized.
  For $||\bm{\theta}^{task}-\bm{\theta}_{t}||_{2}$, under the same conditions of minimizing $\bm{f}^{style}$ or maximizing $\bm{f}^{style}$, the muscle length velocity was correctly modified by velocity maximization and minimization, respectively.
  Note that, at the time of demonstration, the antagonism increases as $\bm{f}^{style}$ is increased, and the joint angle velocity is reduced with high $\bm{f}^{style}$ compared to with low $\bm{f}^{style}$.
}%
{%
  まず, 1自由度関節の半径$\bm{r}$, 動作の筋張力$\bm{f}^{style}$を変化させて実験を行った.
  $r=\{0.03, 0.035, 0.04\}$ [m], $\bm{f}^{style}=\{10, 50, 100, 150, 200\}$ [N]の15の組み合わせについてそれぞれ約60ステップのデータを取得し, これを用いてRNNPBを学習させた.
  訓練時に得られた$\bm{p}_{k}$をPCAを通して2次元で表示したものを\figref{figure:1dofsim-exp1-pb}に示す.
  $r$と$\bm{f}^{style}$の大小に沿って, $\bm{p}_{k}$が綺麗に整列していることがわかる.
  シミュレータを$r=0.04$で固定し, $\bm{p}=\bm{0}$の状態で一度動作してデータ$D$を取得して, 追加の制約からオフラインで更新した$\bm{p}$の値も同様に\figref{figure:1dofsim-exp1-pb}に示す.
  ここでは, $r$の変化が\secref{subsec:update}の(A)に, $\bm{f}^{style}$の変化が(B)に対応するため, (A)の合致制約を入れるか否か, (B)の制約について最大化するか・最小化するか・制約を入れないかの組み合わせである, A, B-maximize, B-minimize, AB-maximize, AB-minimizeについて$\bm{p}$を示している.
  $\bm{p}$の更新の際は$\alpha=0.1$ (最大化の際は符号を逆にする), 学習率は0.01, epochは30とし, (B)については\equref{eq:tension}の制約式を用いている.
  また, originalは$\bm{p}=0$を表す(なお, PCAによって変換されるため, $\bm{p}=\bm{0}$がグラフの原点にあるとは限らない).
  (A)を考慮したものは全て, 学習時に得られた$r=0.04$のときの$\bm{p}_{k}$に近い値を取っており, 現在の$\bm{r}$の値を正しく認識できていることがわかる.
  また, (B)を考慮したものは, 最大化であれば$\bm{f}^{style}=200$ [N]の$\bm{p}_{k}$付近に, 最小化であれば$\bm{f}^{style}=10$ [N]の$\bm{p}_{k}$付近に$\bm{p}$が移動していることがわかる.
  ここで得られた$\bm{p}$を使ってタスクを行ったときの$||\bm{\theta}^{task}-\bm{\theta}_{t}||_{2}$, $||\bm{f}_{t}||_{2}$を\figref{figure:1dofsim-exp1-eval}に示す.
  なお, 以降の実験では, オンライン学習実験の間は$\bm{p}$は連続的に更新される一方, \figref{figure:1dofsim-exp1-eval}のような評価実験の際は, 更新する前と後の$\bm{p}$を固定したうえで, online updateを行わない状態で実験することで, 一貫性のあるグラフを得ている.
  $r=0.04$の際の$\bm{p}_{k}$から遠い, originalとB-minimizeの$\bm{p}$では, $\bm{\theta}_{t}$が$\bm{\theta}^{task}$まで到達しておらず, タスク達成に誤差がのってしまっていることがわかる.
  また, (B)をminimizeした$\bm{p}$では$\bm{f}_{t}$が小さく, (B)をmaximizeした$\bm{p}$では$\bm{f}_{t}$が大きく, (B)に対して制約をかけていない$\bm{p}$では, その中間程度の$\bm{f}_{t}$を示していた.
  B-maximizeについては, (A)の制約は入れていないが, 偶然にも$r=0.04$の$\bm{p}_{k}$付近の$\bm{p}$が得られたため, AB-maximizeと同じような挙動になっている.

  次に, $r=0.04$で固定し, 動作の筋張力$\bm{f}^{style}$と, 動作速度を表す係数$\beta$を変化させて実験を行った.
  $\bm{f}^{style}=\{10, 50, 100\}$ [N], $\beta=\{0.05, 0.10, 0.15\}$の9の組み合わせについてそれぞれ約60ステップのデータを取得し, これを用いてRNNPBを学習させた.
  訓練時に得られた$\bm{p}_{k}$をPCAを通して2次元で表示したものを\figref{figure:1dofsim-exp2-pb}に示す.
  $\bm{f}^{style}$と$\beta$の大小に沿って, $\bm{p}_{k}$が綺麗に整列していることがわかる.
  先の実験と同様に追加の制約からオフラインで更新した$\bm{p}$の値も\figref{figure:1dofsim-exp2-pb}に示す.
  ここでは, $\bm{f}^{style}$と$\beta$について, \equref{eq:tension}と\equref{eq:lvelocity}の制約式を用いて, それぞれについて最大化, 最小化を行う4つの組み合わせについて実験を行った.
  $\bm{p}$の更新の際は$(\alpha_1, \alpha_2)=\{(0.02, 0.3)$ (なお, $\bm{f}^{style}$に関する重みを$\alpha_{1}$, $\beta$に関する重みを$\alpha_{2}$とする), 学習率は0.01, epochは30とした.
  なお, 以降の実験では基本的に(B)の制約についてのみ考えるが, \equref{eq:loss}の(A)の制約式も含めて$\bm{p}$を更新している.
  $\bm{f}^{style}$の方向については最小化・最大化によって比較的意図した通りの方向に$\bm{p}$が更新された.
  それに対して, $\beta$の方向については, 一部意図した通りの方向に$\bm{p}$は更新されていない.
  最適化ごとに$\alpha_1$と$\alpha_2$を調整することで所望の結果を得ることはできるものの, 2つ以上の(B)の制約を考慮するためには, 注意深い重みの比率の調整が必要なことがわかった.
  得られた$\bm{p}$を使ってタスクを行ったときの$||\bm{\theta}^{task}-\bm{\theta}_{t}||_{2}$, $||\bm{f}_{t}||_{2}$を\figref{figure:1dofsim-exp2-eval}に示す.
  \figref{figure:1dofsim-exp2-pb}における$\bm{f}^{style}$の軸に関する$\bm{p}$の配置結果と同じく, $||\bm{f}_{t}||_{2}$はほとんどの場合において最小化・最大化によって意図した通りに制御できており, $\bm{f}^{style}$と$\beta$を両方最小化した場合はのみ$\bm{f}^{style}$が最小化し切れていなかった.
  $||\bm{\theta}^{task}-\bm{\theta}_{t}||_{2}$については, $\bm{f}^{style}$を最小化した場合, $\bm{f}^{style}$を最大化した場合のそれぞれ同じ条件下では, 速度最大化・最小化によって筋長速度が正しく変更できている.
  なお, もともと$\bm{f}^{style}$を大きくすると拮抗が強まり, $\bm{f}^{style}$が小さい場合と比べると速度は出なくなっている.
}%

\subsection{Box Opening by PR2} \label{subsec:pr2-exp}
\switchlanguage%
{%
  The imitated motion trained with the data of 42 demonstrations is shown in \figref{figure:pr2-exp-behavior}, and the trained $\bm{p}_{k}$ is shown in \figref{figure:pr2-exp-pb}.
  In \figref{figure:pr2-exp-behavior}, the hand withdrew when the box was opened, the hand tried to open it when it was closed, and the task was performed even when disturbed.
  The trajectories of $\bm{p}$ when updated online with (1) maximizing or (2) minimizing the joint angle velocity using \equref{eq:jvelocity} at different box angles are shown in \figref{figure:pr2-exp-pb}.
  By minimization and maximization, we can see that $\bm{p}$ is moving in different directions.
  The transition of $||\bm{\theta}_{t}-\bm{\theta}^{init}||_{2}$ when the task is performed using $\bm{p}$ updated by (1) and (2) is shown in \figref{figure:pr2-exp-eval}.
  Here we show the trajectories (``before-\{1, 2, 3\}'' and ``after-\{1, 2, 3\}'') and their averages (``before-ave'' and ``after-ave'') for the three trials before and after updating $\bm{p}$.
  Note that $||\bm{\theta}_{t}-\bm{\theta}^{init}||_{2}$ does not start from 0 rad, because, as a matter of convenience, we show the open box for 3 seconds and then close it to start the task.
  The task of opening the box was accomplished in all cases.
  In (1), the motion before the update of $\bm{p}$ with stagnation in the middle of execution disappeared, and the robot returned to the initial posture immediately after opening the box after the update of $\bm{p}$.
  In (2), the motion from the first movement to the opening of the box became much slower after the update of $\bm{p}$ than before.
}%
{%
  得られた42試行分のデータを使って学習させた際の動作の様子を\figref{figure:pr2-exp-behavior}に, 訓練された$\bm{p}_{k}$を\figref{figure:pr2-exp-pb}に示す.
  \figref{figure:pr2-exp-behavior}では, 箱を開けると手を引き, 閉じると開けようとする, また, 邪魔をしてもタスクを行うことができた.
  このとき, 別々の箱の角度において, 速度を最大化(1), 最小化(2)する制約を加えて$\bm{p}$をオンライン更新したときの$\bm{p}$の軌跡を\figref{figure:pr2-exp-pb}に示す.
  最小化・最大化によって, 別方向に$\bm{p}$が動いていることがわかる.
  この(1), (2)で更新された$\bm{p}$を使ってタスクを行ったときの$||\bm{\theta}_{t}-\bm{\theta}^{init}||_{2}$の遷移を\figref{figure:pr2-exp-eval}に示す.
  なお, 全ての動作に置いて, 箱を開けるタスクは完遂できていた.
  ここでは, $\bm{p}$を更新する前と後について, 3回の動作の軌道(``before-\{1, 2, 3\}'' and ``after-\{1, 2, 3\}'')と, その平均(``before-ave'' and ``after-ave'')を示している.
  なお, 動作の都合上, 箱を開いた状態を3秒見せてから箱を閉じて動作を始めているため, $||\bm{\theta}_{t}-\bm{\theta}^{init}||_{2}$は0 radからは始まっていない.
  (1)では, $\bm{p}$の更新前の, 途中でゆっくり停滞しながら動く動作が消え, 箱を開けてすぐに初期姿勢に戻るような動作が生成された.
  (2)では, 最初の動き出しから箱を開けるまでの動作が, 更新前に非常に比べてゆっくりになった.
}%

\begin{figure}[t]
  \centering
  \includegraphics[width=1.0\columnwidth]{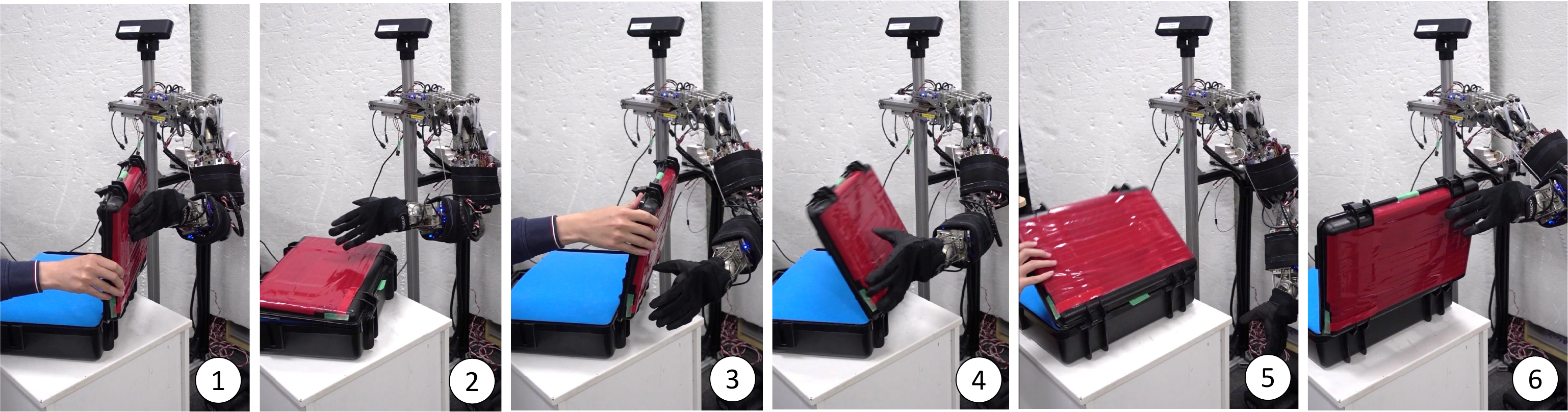}
  \vspace{-3.0ex}
  \caption{The trained behavior of box closing by MusashiLarm.}
  \label{figure:musashilarm-exp-behavior}
  \vspace{-1.0ex}
\end{figure}

\begin{figure}[t]
  \centering
  \includegraphics[width=0.7\columnwidth]{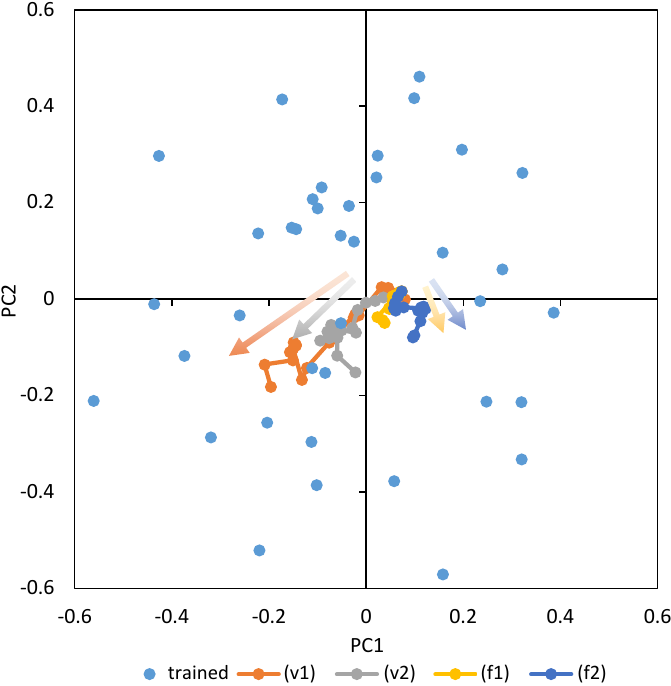}
  \vspace{-1.0ex}
  \caption{Box closing experiment by MusashiLarm: parametric biases trained in data collection phase and their trajectories updated from additional constraints regarding (v1), (v2), (f1), and (f2).}
  \label{figure:musashilarm-exp-pb}
  \vspace{-3.0ex}
\end{figure}

\begin{figure}[t]
  \centering
  \includegraphics[width=1.0\columnwidth]{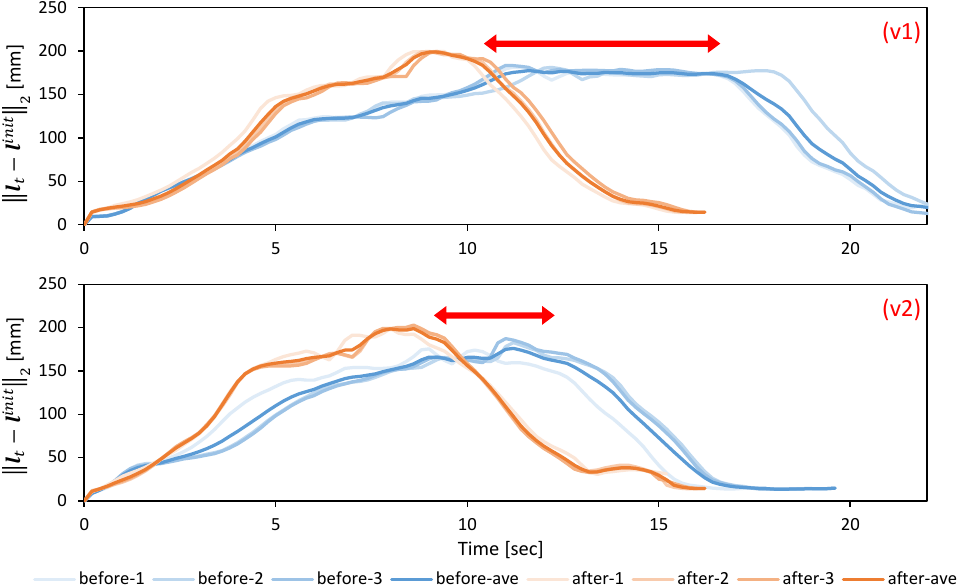}
  \vspace{-3.0ex}
  \caption{Evaluation of box closing experiment by MusashiLarm: transition of $||\bm{l}_{t}-\bm{l}^{init}||_{2}$ when using $\bm{p}$ before and after online update regarding (v1) and (v2).}
  \label{figure:musashilarm-exp-eval-v}
  \vspace{-1.0ex}
\end{figure}

\begin{figure}[t]
  \centering
  \includegraphics[width=1.0\columnwidth]{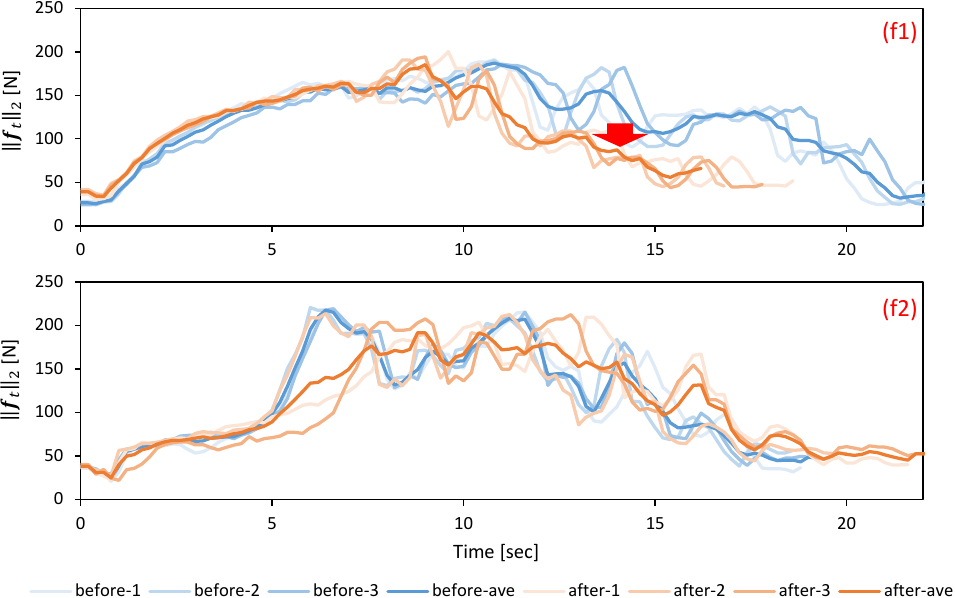}
  \vspace{-3.0ex}
  \caption{Evaluation of box closing experiment by MusashiLarm: transition of $||\bm{f}_{t}||_{2}$ when using $\bm{p}$ before and after online update regarding (f1) and (f2).}
  \label{figure:musashilarm-exp-eval-f}
  \vspace{-3.0ex}
\end{figure}

\subsection{Box Closing by the Musculoskeletal Arm MusashiLarm} \label{subsec:musashilarm-exp}
\switchlanguage%
{%
  The imitated motion trained with the data of 40 demonstrations is shown in \figref{figure:musashilarm-exp-behavior}, and the trained $\bm{p}_{k}$ is shown in \figref{figure:musashilarm-exp-pb}.
  In \figref{figure:musashilarm-exp-behavior}, the hand was withdrawn when the box was closed, the hand tried to close it when it was opened, and the task was performed even when disturbed.
  The trajectories of $\bm{p}$ when updated online with maximizing muscle length velocity (v1, v2) using \equref{eq:lvelocity} and minimizing muscle tension (f1, f2) using \equref{eq:tension} at different box angles (each box angle in experiment (f1), (f2), (v1), and (v2) is different) are shown in \figref{figure:musashilarm-exp-pb}.
  We can see that $\bm{p}$ is moving in the same direction in (v1) and (v2), while (f1) and (f2) do not move much from $\bm{p}=\bm{0}$.
  The transitions of $||\bm{l}_{t}-\bm{l}^{init}||_{2}$ when the task is performed using $\bm{p}$ obtained in (v1) and (v2) are shown in \figref{figure:musashilarm-exp-eval-v}.
  As in \secref{subsec:pr2-exp}, the trajectories of the three trials and their averages are shown for before and after updating $\bm{p}$.
  The task of closing the box was accomplished in all cases.
  We can see that the task execution speed is greatly increased and the constraint is correctly applied.
  The transitions of $||\bm{f}_{t}||_{2}$ when the task is performed using $\bm{p}$ obtained in (f1) and (f2) are shown in \figref{figure:musashilarm-exp-eval-f}.
  In (f1), the muscle tension was changed to a slightly smaller motion than before the update, but in (f2), the result was not so different from before the update.
}%
{%
  得られた40試行分のデータを使って学習させた際の動作の様子を\figref{figure:musashilarm-exp-behavior}に, 訓練された$\bm{p}_{k}$を\figref{figure:musashilarm-exp-pb}に示す.
  \figref{figure:musashilarm-exp-behavior}では, 箱を閉めると手を引き, 開けると閉めようとする, また, 邪魔をしてもタスクを行うことができた.
  このとき, それぞれ別々の箱の角度で速度を最大化(v1, v2), 筋張力を最小化(f1, f2)する制約を加えてオンライン学習させたときの$\bm{p}$の軌跡を\figref{figure:musashilarm-exp-pb}に示す.
  (v1)と(v2)は同じ方向へ$\bm{p}$が動いていることがわかるが, (f1)と(f2)は$\bm{p}=\bm{0}$からそれほど大きく動いていない.
  (v1)と(v2)で得られた$\bm{p}$を使ってタスクを行ったときの$||\bm{l}_{t}-\bm{l}^{init}||_{2}$の遷移を\figref{figure:musashilarm-exp-eval-v}に示す.
  \secref{subsec:pr2-exp}と同様に, $\bm{p}$を更新する前と後について, 3回の動作の軌道と, その平均を示している.
  大きくタスク実行速度が速くなり, 正しく制約が加えられていることがわかる.
  また, (f1)と(f2)で得られた$\bm{p}$を使ってタスクを行ったときの$||\bm{f}_{t}||_{2}$の遷移を\figref{figure:musashilarm-exp-eval-f}に示す.
  (f1)では更新前に比べて筋張力が多少小さな動作に変更されたが, (f2)では更新前とあまり変わらない結果が得られた.
}%

\section{Discussion} \label{sec:discussion}
\switchlanguage%
{%
  First, from the simulated 1-DOF tendon arm, we found that our method can take into account the minimization and maximization constraints on muscle tension and motion speed, as well as the matching constraints on robot configuration such as the joint radius.
  By training RNNPB from the obtained demonstration data, $\bm{p}_{k}$ is neatly self-organized.
  By updating $\bm{p}$ to consider the constraints on the motion style and robot configuration, we can obtain $\bm{p}$ for the intended style and configuration.
  By using the updated $\bm{p}$, the muscle tension and velocity during task execution could be controlled as intended.
  At the same time, it is difficult to adjust the ratio of the weights of loss functions when there are two or more constraints on the motion style.

  Next, the box opening experiment by PR2 shows that our method can be applied to an axis-driven robot with multiple degrees of freedom, and that $\bm{p}$ can be updated not only offline but also online.
  Since $\bm{p}$ is updated in the opposite direction by velocity maximization and minimization, it is considered that the information on the motion style is neatly self-organized.
  By actually executing the task using the obtained $\bm{p}$, we were able to adjust the velocity as intended, demonstrating the effectiveness of this study.

  Finally, the box closing experiment by MusashiLarm shows that our method can be applied to a tendon-driven robot with multiple degrees of freedom.
  The fact that $\bm{p}$ is updated in the same direction by velocity minimization at different box angles suggests that the information on the motion style is neatly self-organized, as in the box-opening experiment by PR2.
  By actually executing the task using the obtained $\bm{p}$, we were able to adjust the velocity as intended, indicating the effectiveness of this study.
  On the other hand, the muscle tension could not be adjusted as intended as the velocity, but this is thought to be because there were many variations in the demonstration for the motion speed, while not much difference was obtained for the muscle tension.
  In this study, we did not explicitly change $\bm{f}^{ref}$, but if we change the target muscle tension from myoelectricity of the human for example, $\bm{f}^{ref}$ will explicitly change and the constraint of the muscle tension can be considered more clearly.
  Although not directly related to the purpose of this study, this is the first time that a tendon-driven system with redundant muscles has been used for imitation learning, and it is expected that imitation learning utilizing a flexible body will be further developed in the future.

  The limitations of this study are described as below.
  First, this method can only add constraints to the motion style within the range of variability obtained during the demonstration.
  Therefore, it is difficult to apply this method to unambiguous tasks that can only be performed in similar styles, and it is not possible to slow down the speed or reduce the power beyond the variation in the demonstration.
  It may be necessary to change a control device to advance a wide variety of actions during the demonstration.
  It is also necessary to verify how the behavior of task accomplishment changes when $\bm{p}$ deviates greatly from $\bm{0}$ due to unreasonable online learning, and to what extent the adaptability to human intervention, hindrance, etc., changes.
  Second, since the loss function cannot be defined for values that are not output from the network, this study cannot be applied to non-quantifiable behavior styles.
  On the other hand, if annotations are made for non-quantifiable behavioral styles, it will be possible to handle such styles by determining the desired non-quantifiable behavioral style and using parametric bias around it.
  Third, since the motion style of a single demonstration is embedded in a single parametric bias, it is not possible to consider the case where the style changes during the demonstration.
  For more complex behaviors, it is necessary to consider how to embed parametric bias when training and how to divide the demonstration data.

  In this study, we mainly focused on the constraints of speed and force, but in the future, we would like to consider more constraints.
  For example, if an auditory sensor is added, we can define a loss function to minimize the sound during the motion, and if a contact sensor is added, we can limit when the hand makes contact with the environment and change the applied force.
  We would like to aim at imitation learning using more multimodal sensors.
  In addition, the space of parametric bias needs to be examined in more detail in the future.
  It is necessary to further verify how multiple motion styles are embedded depending on the number of dimensions of parametric bias.
  Finally, although we have dealt with the matching constraint of robot configuration of the joint radius in the simulation, we would like to discuss whether it is possible to consider the case where the body changes due to muscle rupture or where the friction of the body increases due to deterioration over time.
  Although not regarding imitation learning, we have been able to capture differences in robot calibration and grasping tools by parametric bias \cite{kawaharazuka2020dynamics}, and so we would like to verify whether it is possible to take into account large changes in robot configuration, movement pattern, etc., online.
}%
{%
  まず, 1自由度腱駆動ロボットシミュレーションから, 本手法によって, 筋張力や動作速度に関する最小化・最大化制約を考慮できること, また, 関節半径等のrobot configurationの合致制約等も同時に考慮することが可能なことがわかった.
  得られたdemonstrationのデータからRNNPBを学習させることで, $\bm{p}_{k}$が綺麗に自己組織化される.
  動作スタイル・ロボット状態に関する制約を考慮して$\bm{p}$更新することで, 意図した動作スタイル・ロボット状態の$\bm{p}$を求めることができる.
  実際にこの更新された$\bm{p}$を用いることで, タスク実行時の筋張力・速度を意図した通りに制御できることがわかった.
  同時に, 動作スタイルに関する制約が2つ以上ある場合は, それらの重みの比率の調整が難しいということもわかった.

  次に, PR2による箱開け動作により, 多自由度の軸駆動型ロボットにおいても本研究が適用可能であること, $\bm{p}$はオフラインだけでなくオンラインでも更新可能であることがわかった.
  速度最大化・最小化によって$\bm{p}$が逆方向に更新されることから, それら動作スタイルに関する情報が綺麗に自己組織化されていると考えられる.
  実際に得られた$\bm{p}$を使ってタスクを実行することで, 速度が意図したように調整出来ており, 本研究の有効性が示された.

  最後に, MusashiLarmによる箱閉め実験により, 多自由度の腱駆動型ロボットにおいても本研究が適用可能であることがわかった.
  異なる箱の角度において, 速度最小化により$\bm{p}$が同じ方向に更新されることから, PR2の箱開け実験と同様に, 動作スタイルに関する情報が綺麗に自己組織化されていると考えられる.
  実際に得られた$\bm{p}$を使ってタスクを実行することで, 速度が意図したように調整出来ており, 本研究の有効性が示された.
  一方, 筋張力に関しては速度ほど意図した調整ができなかったが, これはデモンストレーション内で, 動作速度に関しては様々なバリエーションがあったのに対して, 筋張力についてはそれほど違いが得られなかったからだと考えられる.
  本研究では明示的に$\bm{f}^{ref}$について変化を加えなかったが, 例えば人間の筋電からロボットの筋張力入力を変化させるようにすると, 明示的に$\bm{f}^{ref}$が変化し, より明確に筋張力の制約を考慮可能になると考えられる.
  また, 本研究の目的とは直接は関係ないが, 冗長な筋を持つ腱駆動型による模倣学習は初めてであり, 柔軟な身体を活かした模倣学習が今後より期待される.

  本研究のlimitationについて述べる.
  まず, 本研究はあくまでデモンストレーション中に得られたばらつきの範囲内でしか動作スタイルに制約を加えることができない.
  そのため, ほとんど同じスタイルでしか実行できないような曖昧性のないタスクへの適用は難しく, また, デモンストレーションのばらつき以上に速度を遅くしたり, 力を弱めたり等はできない.
  デモンストレーション時にバラエティーに富んど動作を行うような操作デバイス側の工夫等が必要な可能性がある.
  また, 無理なオンライン学習により$\bm{p}$が$\bm{0}$から大きく外れた場合, 以下にタスク達成の挙動が変化するのか, human interventionやhinderance等による適応性はどの程度変化するのかについても検証が必要である.
  次に, 本研究の形では, ネットワークから出力されない値にはcost functionを定義できないので, non-quantifiable behavior styleに適用できない点が挙げられる.
  一方, styleに対して定量化できないbehavior styleに関するannotationが行われていれば, desired non-quantifiable behavior styleを決め, その周辺のparametric biasを使うということによって, 扱うことは可能になると考えている.
  最後に, 一回のデモンストレーション時の動作スタイルを一つのparametric biasに埋め込むため, 途中でスタイルが変わる場合等を考慮できないという点である.
  今後より複雑な動作を実行する際は, 学習時におけるparametric biasの入れ方, 動作データの分割等について, 考察していく必要がある.

  本研究では主に速度と力の動作スタイル制約について扱ったが, 今後, さらに多くの制約を考慮していきたい.
  例えば, 聴覚センサを加えれば, 動作中の音をなるべく小さくする等の制約を考慮でき, 接触センサを加えれば, いつ環境と接触するのかを制限したり, 力の書け具合を変えることができるはずである.
  より多数のマルチモーダルなセンサを利用した模倣学習を目指したい.
  また, parametric biasの空間についても今後より詳細な検証が必要であり, 複数のmotion styleが, parametric biasの次元数によって如何に埋め込まれるのか, それらの構造についてさらなる検証をしていく必要がある.
  最後に, シミュレーションにおいて関節半径のrobot configurationの合致制約について扱ったが, 今後は, 筋が一本切れて身体が変化する場合や, 経年劣化により身体の摩擦が増える場合等についても考えていきたい.
  これまで, ロボットのcalibrationが変化した場合, 把持する道具が異なる場合について, 模倣学習ではないもののparametric biasによりそれらの違いをcaptureできているため\cite{kawaharazuka2020dynamics}, robot configuration, movement pattern等の大きな変化もオンラインで考慮可能であるかどうか検証していきたい.
}%

\section{Conclusion} \label{sec:conclusion}
\switchlanguage%
{%
  In this study, we proposed an imitation learning method that can take into account additional constraints on the motion style in addition to imitating the demonstrated task.
  The variability of the human demonstration is embedded into parametric bias.
  By defining loss functions for additional constraints on predictive states and control variables and updating the parametric bias offline or online, additional constraints can be considered.
  We have successfully applied this study to PR2 and the musculoskeletal humanoid MusashiLarm, and have succeeded in minimizing muscle tension and changing muscle length and joint velocity.
  In the future, we would like to consider more complex additional constraints on sound, contact force, obstacle avoidance, etc., and longer term task execution.
}%
{%
  本研究では, 学習したtaskを模倣する以外に追加の動作スタイル制約を考慮可能な模倣学習手法の提案を行った.
  人間の教示による動作のばらつきをparametric biasに埋め込んだ.
  予測状態や制御変数に関する追加制約の損失関数を定義し, このparametric biasをオフラインまたはオンラインで更新することで, 追加制約を考慮することが可能となった.
  本研究を柔軟で冗長かつ多数のセンサをを持つ筋骨格ロボットやPR2に適用し, 筋張力最小化や筋長・関節速度変化等を行うことに成功した.
  今後は, 動作中の音や接触力, 障害物回避等のより複雑な追加制約, より長期的なタスク実行についても考えていきたい.
}%

{
  \bibliographystyle{IEEEtran}
  \bibliography{main}
}

\end{document}